\documentclass[twoside]{article}

\usepackage[accepted]{aistats2024}

\usepackage[T1]{fontenc}
\usepackage[utf8]{inputenc}

\usepackage[round]{natbib}

\usepackage{hyperref}
\hypersetup{colorlinks=false, hidelinks}

\usepackage{amsmath, amssymb, bm, mathtools}

\usepackage{booktabs}
\usepackage{multirow}
\usepackage{xcolor}

\usepackage[lined,ruled,vlined,algo2e]{algorithm2e}
\SetKw{KwForIn}{in}
\SetKw{KwUntil}{until}
\SetKwInput{KwInputs}{Inputs}
\SetKwInput{KwOutputs}{Outputs}

\newcommand{\cu}[1]{
	\ifcat\noexpand#1\relax
	\bm{#1}
	\else
	\mathbf{#1}
	\fi
}

\newcommand{\diff}{\mathop{}\!\mathrm{d}}

\newcommand{\expp}{\mathrm{e}}

\newcommand{\cond}{{\;|\;}}

\let\lim\relax
\DeclareMathOperator*{\argmin}{arg\,min\,}  % Argmin
\DeclareMathOperator*{\lim}{lim\,}  % inf
\let\grad\relax
\DeclareMathOperator*{\grad}{\nabla\!}

\newcommand{\expecsym}{\operatorname{\mathbb{E}}}     % Expec
\newcommand{\covsym}{\operatorname{Cov}}     % Covariance
\newcommand{\varrsym}{\operatorname{Var}}     % Variance
\newcommand{\diagsym}{\operatorname{diag}}     % Diagonal matrix
\newcommand{\tracesym}{\operatorname{tr}}           % Trace

\let\expec\relax
\let\cov\relax
\let\varr\relax
\let\diag\relax
\let\trace\relax

\makeatletter
\newcommand{\expec}{\@ifstar{\@expecauto}{\@expecnoauto}}
\newcommand{\@expecauto}[1]{\expecsym \left[ #1 \right]}
\newcommand{\@expecnoauto}[1]{\expecsym [#1]}
\newcommand{\expecbig}[1]{\expecsym \bigl[ #1 \bigr]}

\newcommand{\expecbigg}[1]{\expecsym \biggl[ #1 \biggr]}

\newcommand{\cov}{\@ifstar{\@covauto}{\@covnoauto}}
\newcommand{\@covauto}[1]{\covsym \left[ #1 \right]}
\newcommand{\@covnoauto}[1]{\covsym [#1]}

\newcommand{\varr}{\@ifstar{\@varrauto}{\@varrnoauto}}
\newcommand{\@varrauto}[1]{\varrsym \left[ #1 \right]}
\newcommand{\@varrnoauto}[1]{\varrsym [#1]}

\newcommand{\diag}{\@ifstar{\@diagauto}{\@diagnoauto}}
\newcommand{\@diagauto}[1]{\diagsym \left( #1 \right)}
\newcommand{\@diagnoauto}[1]{\diagsym (#1)}

\newcommand{\trace}{\@ifstar{\@traceauto}{\@tracenoauto}}
\newcommand{\@traceauto}[1]{\tracesym \left( #1 \right)}
\newcommand{\@tracenoauto}[1]{\tracesym (#1)}

\makeatother

\newcommand*{\R}{\mathbb{R}} % Set of real numbers
\makeatletter
\@ifundefined{thmenvcounter}{}
{%

}
\makeatother

\begin{document}

\twocolumn[

\aistatstitle{On Feynman--Kac training of partial Bayesian neural networks}

\aistatsauthor{ Zheng Zhao \And Sebastian Mair \And Thomas B. Sch\"{o}n \And Jens Sj\"{o}lund }

\aistatsaddress{ Uppsala University, Sweden } ]

\begin{abstract}
Recently, partial Bayesian neural networks (pBNNs), which only consider a subset of the parameters to be stochastic, were shown to perform competitively with full Bayesian neural networks. 
However, pBNNs are often multi-modal in the latent variable space and thus challenging to approximate with parametric models. 
To address this problem, we propose an efficient sampling-based training strategy, wherein the training of a pBNN is formulated as simulating a Feynman--Kac model.
We then describe variations of sequential Monte Carlo samplers that allow us to simultaneously estimate the parameters and the latent posterior distribution of this model at a tractable computational cost. 
Using various synthetic and real-world datasets we show that our proposed training scheme outperforms the state of the art in terms of predictive performance.\looseness=-1
\end{abstract}

\section{INTRODUCTION}
\label{sec:intro}
Bayesian neural networks (BNNs) are an important class of machine learning models for quantifying uncertainty. 
However, computing BNN posterior distributions is an open challenge~\citep{izmailov2021bayesian} since many standard statistical inference methods such as Markov chain Monte Carlo (MCMC) are poorly suited to the combination of a high-dimensional parameter space and a massive number of data points~\citep{Papamarkou2022}. 
Many approximate solutions have been proposed to overcome this computational challenge, for example, MCMC with stochastic likelihoods~\citep{Andrieu2009, Welling2011, ChenTQ2014, ZhangRuqi2020}, parametric approximations of the posterior distributions in the form of variational Bayes~\citep{Blundell2015, Gal2016} or stochastic averaging methods~\citep{izmailov2018averaging, maddox2019simple}. In a similar vein, \citet{izmailov2020subspace} conduct variational inference in a subspace of the parameter space. Other ways include, for instance, restricting the inference only on the last layer~\citep{ober2019benchmarking, kristiadi2020being}. 

In this paper, we focus on the training of partial Bayesian neural networks (pBNNs), a family of BNNs where only part of the model parameters are random variables.
This is motivated by the recent work of, for example, \cite{daxberger2021bayesian}~and~\citet{Sharma2023}, which show that pBNNs and standard BNNs are comparable in terms of prediction performance despite the dimensionality of the random variable in the pBNNs being much smaller than standard BNNs. While both approaches make posterior inference much easier, it comes at the cost of making the pBNN a latent variable model that requires estimation of both the deterministic parameters and the posterior. 

To be more precise, let $x\mapsto f(x; \phi, \psi)$ be a neural network parametrised by a deterministic parameter $\psi\in\R^w$ and a random variable $\phi\in\R^d$ that follows a given prior distribution $\pi(\phi)$. 
Suppose that we have a dataset $\lbrace (x_n, y_n) \rbrace_{n=1}^N$ with an associated likelihood conditional density function $p(y_n \cond \phi; \psi)$\footnote{Throughout the paper we omit the data covariate $x_n$ of $y_n$ to make the notation cleaner.}, and that the observation random variables are independent conditioned on $\phi$. 
The goal of training is twofold. First, to learn the deterministic parameter $\psi$ from the dataset, and second, to compute the posterior distribution $p(\phi \cond y_{1:N} ; \psi)$, where $y_{1:N} \coloneqq \lbrace y_1, y_2, \ldots, y_N \rbrace$. This is a classical parameter estimation problem in a latent variable model~\citep{Cappe2005}.
The main benefit of pBNNs is that when the dimension $d \ll w$ is relatively small, the second goal of computing the posterior distribution with fixed $\psi$ is generally tractable. However, the first goal of estimating the parameter $\psi$ remains to be solved. 

For practitioners, a popular approach is to jointly optimise $\phi$ and $\psi$ according to the maximum a posteriori (MAP) objective $(\phi, \psi) \mapsto \log p(y_{1:N} \cond \phi; \psi) + \log p(\phi)$, see, for example, \citet{daxberger2021bayesian} and~\citet{Sharma2023}. 
Then, the estimated $\psi$ is used to compute the posterior distribution over $\phi$ by any applicable Bayesian inference method (e.g., MCMC).
The advantages of this method lie in the simplicity and fast computation. 
Moreover, there is a plethora of optimisers (e.g., stochastic gradient descent and its variants) specifically designed for the setting of large $w$ and $N$. 
However, for multi-modal distributions the MAP estimated $\psi$ and $\phi$ are prone to be restricted on a single mode. \looseness=-1

Another popular approach is to use the maximum likelihood estimation (MLE) for estimating $\psi$. 
Compared to the aforementioned MAP approach, the MLE objective takes all possible outcomes of the random variable $\phi$ into account by integrating out $\phi$, and the estimator can be consistent~\citep{Cappe2005}. 
However, the MLE method is not directly applicable to pBNNs since the marginal likelihood $p(y_{1:N}; \psi) = \int p(y_{1:N} \cond \phi; \psi) \, \pi(\phi) \diff \phi$ is in general intractable.
Instead, various lower bounds on the marginal likelihood are used by, for instance, brute-force Monte Carlo, expectation maximisation~\citep[EM,][]{Bishop2006Book}, and variational Bayes~\citep[VB,][]{blei2017variational,sjolund2023tutorial}. 
However, EM and VB often compromise in using parametric families of distributions, and they often incur Monte Carlo computation overheads to approximate expectations. 
Another common approach is to transform the (gradient of) MLE into an expectation with respect to the posterior distribution by Fisher's identity~\citep[see, e.g.,][Chap. 10]{Cappe2005}, and then use an MCMC sampler to approximate the expectation (at the cost of demanding computations).

In this paper, we study how to apply and adapt sequential Monte Carlo (SMC) methods to train pBNNs, by representing the training as a simulation of a Feynman--Kac model~\citep{ChopinBook2020}. 
More specifically, the Feynman--Kac model is composed of a sequence of potential functions given by the likelihood model, and a sequence of invariant Markov kernels that anneals to the target posterior distribution $p(\phi \cond y_{1:N}; \psi)$. 
Computing the target posterior distribution then amounts to a sequential simulation of the Feynman--Kac model, and SMC is the natural sampling framework for the model. 
The motivation for studying SMC samplers to train pBNNs is that SMC samplers are able to simultaneously produce consistent MLEs and to sample from the posterior distribution while retaining a tractable computational cost. 
Compared to most MCMC-based methods, SMC samplers are immediately parallelisable~\citep{Leee2010} by leveraging graphics processing units, and are easier to calibrate. On the other hand, parallelising MCMC is still an open problem~\citep[see, e.g.,][]{Jacob2020}.

Our contributions are as follows. (i) We show and discuss how to apply and adapt SMC samplers to train pBNNs (Section~\ref{sec:feynman-kac-smc}). (ii) We propose approximate SMC samplers that are scalable in the number of data points and are therefore better suited to pBNNs (Section~\ref{sec:scalable-smcs}). (iii) We benchmark the proposed samplers on a variety of synthetic and real-world datasets, and the results show that the proposed training scheme is state-of-the-art in terms of predictive performance (Section~\ref{sec:experiments}). 

\section{TRAINING VIA FEYNMAN--KAC}
\label{sec:feynman-kac-smc}
Recall that we aim to find a maximum likelihood estimate for the deterministic part of the pBNN and then to compute the posterior distribution $p(\phi \cond y_{1:N}; \psi)$ of the stochastic part of the pBNN based on the learnt parameter $\psi$. In this section, we recap a sequential online algorithm to recursively compute the target posterior distribution via a Feynman--Kac model and how to jointly estimate the parameter $\psi$.

To ease the exposition of the idea, let us for now suppose that the deterministic variable $\psi$ in the pBNN is fixed. Due to the conditional independence of the observations, the posterior distribution admits a recursion
\looseness=-1
\begin{equation}
	p(\phi \cond y_{1:n}; \psi) = \frac{p(y_n \cond \phi; \psi)}{z_n(\psi)} \, p(\phi \cond y_{1:n-1}; \psi),
	\label{equ:radon-nikodym-ratio}
\end{equation}
for $n=1, 2, \ldots, N$, where $z_n(\psi) \coloneqq p(y_n \cond y_{1:n-1}; \psi)$ is the normalising constant, and the initial is defined by $p(\phi \cond y_{1:0}; \psi) \coloneqq \pi(\phi)$. If we have computed the distribution $p(\phi \cond y_{1:n-1}; \psi)$ for some $n$, we can then compute the next $p(\phi \cond y_{1:n}; \psi)$ by Equation~\eqref{equ:radon-nikodym-ratio}. Ultimately, we continue the iteration until $n=N$ to reach the target posterior distribution. This recursion is the gist of sequential online learning or annealing. 

As in most Bayesian inference problems, the challenge lies in the computationally intractable normalising constant $z_n(\psi)$. Hence, in practice, we often use a tractable sequence of approximations $Q_N \coloneqq \lbrace q_n(\phi \cond y_{1:n}; \psi) \colon n=1,2,\ldots, N\rbrace$ such that $q_n(\phi \cond y_{1:n}; \psi) \propto q_{n-1}(\phi \cond y_{1:n-1}; \psi)$ is computable for all $n$'s. 
A convenient choice is a Gaussian $q_n$ and then to use, for example, Taylor expansions, variational Bayes~\citep{Opper1999}, or Gauss quadrature methods~\citep{Golub2010} to run the sequence. 
However, in the context of BNNs, Gaussian approximations can result in large errors~\citep{foong2020expressiveness} particularly when the true posterior distribution is non-Gaussian (e.g., multi-modal ones). This motivates us to use a Monte Carlo (MC)-based method to come up with such an approximate sequence.

\subsection{Sequential Monte Carlo sampling}
\label{sec:smc}
Sequential Monte Carlo (SMC) samplers~\citep{DelMoral2006, ChopinBook2020} are natural Monte Carlo (MC) methods for approximating the target posterior distribution in the sequential learning framework. Specifically, we choose $Q_N = \lbrace S_n^J \colon n=1,2,\ldots, N \rbrace$, where $S_n^J \coloneqq \lbrace (w_{n, j}, \phi_{n, j}) \colon j = 1,2,\ldots, J\rbrace$ are $J$ weighted Monte Carlo samples that approximately represent $p(\phi \cond y_{1:n}; \psi)$. Suppose that we are able to draw the initial samples $S_{0}^J \sim \pi(\phi)$ from the given prior, then we can recursively compute $S_n^J$ for any $n=1,2,\ldots, N$ via
\begin{align}
	\phi_{n, j} &= \phi_{n-1, j}, \label{equ:sis}\\
	\overline{w}_{n, j} &= w_{n-1, j} \, p(y_n \cond \phi_{n-1, j}; \psi), \nonumber
\end{align}
and normalisation $w_{n, j} = \overline{w}_{n, j} \, / \, \sum_{i=1}^J \overline{w}_{n, i}$ for all samples $j=1,2,\ldots,J$. This approximation is consistent in the sense that the resulting Dirac measure given by $S_n^J$ converges weakly to that of $p(\phi \cond y_{1:n}; \psi)$ as $J\to\infty$ for all $n$~\citep{ChopinBook2020}. Furthermore, the method is particularly favourable in the pBNN context, as the dimension $d$ for the stochastic part of the pBNN is usually not large which in turn allows us to use significantly more MC samples compared to full BNNs. This method dates back to~\citet{Neal2001} who simulates static target distributions with a likelihood tempering~\citep[see also][]{Chopin2002}. 

However, the SMC chain using Equation~\eqref{equ:sis} rarely works in reality, since the samples $\lbrace \phi_{n, j} \rbrace_{j=1}^J$ will become less informative and the weights $\lbrace w_{n, j} \rbrace_{j=1}^J$ will degenerate as $n$ increases~\citep{DelMoral2006}. In practice, we additionally introduce a Markov transition kernel $h_n(\cdot \cond \phi_{n-1})$ between each step $(n-1, n)$, so that the samples $\lbrace \phi_{n, j} \rbrace_{j=1}^J$ are perturbed according to the kernel. Specifically, the update of samples in Equation~\eqref{equ:sis} modifies to
\begin{equation}
	\phi_{n, j} \cond \phi_{n-1, j} \sim h_n(\phi_{n, j} \cond \phi_{n, j-1}).
	\label{equ:sis-markov}
\end{equation}
Then, we arrive at the so-called (marginal) Feynman--Kac model
\begin{equation}
	\begin{split}
		&q_N(\phi_N \cond y_{1:N}; \psi) \\
		&\coloneqq \frac{1}{l_N(\psi)}\int \prod_{n=1}^N p(y_n \cond \phi_n; \psi) \prod _{n=1}^N h_n(\phi_n \cond \phi_{n-1}) \\
		&\qquad\qquad\quad\times \pi(\phi_0) \diff \phi_{0:N-1}, 
	\end{split}
	\label{equ:feynman-kac}
\end{equation}
where $l_N(\psi)$ is the normalising constant~\citep[see the definition of Feynman--Kac in][Chap. 5]{ChopinBook2020}. Now, to guarantee that the terminal $q_N(\phi_N \cond y_{1:N}; \psi)$ exactly hits the target posterior distribution $p(\phi \cond y_{1:N}; \psi)$, we need to choose the Markov kernel in a way that $h_n$ leaves the previous posterior distribution $p(\phi \cond y_{1:n-1}; \psi)$ invariant, viz., $\int h_n(\phi \cond \phi_{n-1}) \, \rho(\phi_{n-1}) \diff \phi_{n-1} = p(\phi \cond y_{1:n-1}; \psi)$ for any distribution $\rho$. Note that $h_n$ indeed depends on $y_{1:n-1}$ and $\psi$, but we omit them for clean notation. This choice of Markov kernel also ensures that the marginal likelihood is given by $l_N(\psi) = \prod_{n=1}^N z_n(\psi) = p(y_{1:N}; \psi)$.

It is now clear that computing the target posterior distribution $p(\phi \cond y_{1:N}; \psi)$ amounts to simulating the Feynman--Kac model in Equation~\eqref{equ:feynman-kac}. The SMC sampler simulates the Feynman--Kac model with the weighted samples $S_n^J$ for $n=0,1,\ldots,N$ using Equations~\eqref{equ:sis} and~\eqref{equ:sis-markov}. Moreover, recall that we need to estimate the parameter $\psi$ via MLE
\begin{equation*}
	\argmin_{\psi \in\R^w} -\log l_N(\psi).
\end{equation*}
It turns out that the SMC sampler produces a consistent estimate $z_n(\psi) \approx \sum_{j=1}^J \overline{w}_{n, j}$, so that we can estimate the log-likelihood via $\log l_N(\psi) \approx \sum_{n=1}^N \log \bigl(\sum_{j=1}^J \overline{w}_{n, j} \bigr)$ as a by-product of the computation of the posterior distributions. Then, we can solve the optimisation problem above by using any optimiser for training neural networks. For a detailed exposition of SMC parameter estimators, see, for example, \citet{Johansen2008}, \citet{Schon2011}, and~\citet{Kantas2015}. We summarise the SMC training of pBNNs in Algorithm~\ref{alg:standard-smc} together with a gradient descent-based optimisation of the MLE objective. Within the algorithm, we use the shorthand $\ell_n(\cdot)$ for the approximation to the marginal log-likelihood $\log l_n(\cdot)$, and we also use $r(i)$ for the learning rate at the $i$-th iteration.

\begin{algorithm2e}[h]
	\SetAlgoLined
	\DontPrintSemicolon
	\KwInputs{Training data $\lbrace (x_n, y_n) \rbrace_{n=1}^N$, number of samples $J$, initial parameter $\psi_0$, learning rate function $r$}
	\KwOutputs{The MLE estimate $\psi_i$ and weighted posterior samples $\lbrace (w_{N, j}, \phi_{N, j}) \rbrace_{j=1}^J \sim p(\phi \cond y_{1:N}; \psi_i)$}
	\For{$i=1,2,\ldots$ \KwUntil convergent}{%
		Draw $\lbrace \phi_{0, j} \rbrace_{j=1}^J \sim \pi(\phi)$\;
		$w_{0, j} = 1 \, / \, J$ for all $j=1,2,\ldots,J$\;
		$\ell_0(\psi_{i-1}) = 0$\;
		\For(\tcp*[f]{Parallelise $j$}){$n=1$ \KwTo $N$}{%
			Resample $\lbrace (w_{n-1, j}, \phi_{n-1, j})\rbrace_{j=1}^J$ if needed\;
			Draw $\phi_{n, j} \cond \phi_{n-1, j} \sim h_n(\phi_{n, j} \cond \phi_{n, j-1})$\;
			$\overline{w}_{n, j} = w_{n-1, j} \, p(y_n \cond \phi_{n, j}; \psi_{i-1})$\;
			$\ell_n(\psi_{i-1}) = \ell_{n-1}(\psi_{i-1}) - \log\bigl(\sum_{j=1}^J \overline{w}_{n, j}\bigr)$\;
			$w_{n, j} = \overline{w}_{n, j} \, / \, \sum_{k=1}^J \overline{w}_{n, k}$\;
		}
		$\psi_i = \psi_{i-1} + r(i) \grad\ell_N(\psi_{i-1})$
	}
	\caption{Sequential Monte Carlo (SMC) sampler for pBNN}
	\label{alg:standard-smc}
\end{algorithm2e}

\section{SCALABLE SEQUENTIAL MONTE CARLO SAMPLERS}
\label{sec:scalable-smcs}
We still have a few critical challenges left to solve in order to apply the SMC sampler in Algorithm~\ref{alg:standard-smc} for training pBNNs. First, the gradient computation $\grad\ell_N$ might be biased due to the non-differentiability of the resampling and Markov transition steps. Second, it is not easy to design the Markov kernel $h_n$ and also to compute the Markov moves. Third, the algorithm does not scale well in the number of data points $N$. In particular, for each gradient descent step to compute $\grad \ell_N$, we need a complete SMC loop over the entire dataset. In what follows, we detail these issues and show how to tackle them.

\subsection{Compensating gradient biases}
\label{sec:gradient-biases}
The non-differentiability is a notorious issue for sequential Monte Carlo samplers, since most resampling techniques (e.g., systematic and stratified) and MCMC algorithms incur discrete randomness. This in turns induces biases in the gradient computation $\grad \ell_N$. There are recent developments to tackle the differentiability issue by, for example, optimal transport-based smooth resampling~\citep{Corenflos2011} and coupled MCMC~\citep{Gaurav2022, Gaurav2023}, but they come at the cost of introducing additional computation costs and calibrations. However, we can in fact avoid differentiating the SMC algorithm by invoking Fisher's identity~\citep[see, e.g.,][Chap. 10]{Cappe2005}, which states that $\grad \log l_N(\psi) = \int \grad \log p(y_{1:N}, \phi; \psi) \, p(\phi \cond y_{1:N}; \psi) \diff \phi$. Moreover, unlike particle filtering in the system identification context~\citep[see, e.g.,][Sec. 2.2]{Poyiadjis2011}, the model design facilitates $\grad \log p(y_{1:N}, \phi; \psi) = \grad \log p(y_{1:N} \cond \phi; \psi)$. Hence, we can modify the gradient computation in Algorithm~\ref{alg:standard-smc} to
\begin{equation}
	\grad \ell_N(\psi) = \sum_{j=1}^J w_{N, j}\grad \log p(y_{1:N} \cond \phi_{N, j}; \psi),
\end{equation}
which only requires the log-likelihood to be differentiated. \looseness=-1

\subsection{Choosing the Markov kernel}
\label{sec:markov-kernel}
Recall the definition of the Markov kernel $h_n$: It leaves the posterior distribution $p(\phi \cond y_{1:n-1}; \psi)$ invariant. Since the joint $p(y_{1:n-1} \cond \phi; \psi) \, \pi(\phi)$ is usually analytically available, it is common to use any MCMC chain for $h_n$~\citep{DelMoral2006}. In particular, when the problem dimension $d$ is low, using a standard random walk Metropolis--Rosenbluth--Teller--Hasting MCMC suffices in practice. If the dimension is relatively high, we can also leverage the gradient information of the energy to define the kernel with, for instance, a Langevin dynamic. However, it is worth remarking that choosing the number of MCMC steps is non-trivial, and this still remains an open challenge in the SMC community, see, for example, the discussion in~\citet[][Sec. 17.2]{ChopinBook2020} or a recent development in~\citet{Dau2021}.
\looseness=-1

The computational problem of using MCMC for the kernels is that at each $n$, we need to load all the data before $n$ to evaluate the energy function. For gradient-based MCMCs, we also need to compute $\grad_\phi \log p(\phi \cond y_{1:n-1}; \psi)$. This computation becomes even more demanding when $n$  approaches $N$ (which is large), and hence the computational cost grows at least quadratically in $n$. A straightforward remedy to this problem is to use pseudo-marginal MCMC samplers~\citep[see, e.g.,][]{Andrieu2009, Welling2011, ChenTQ2014, ZhangRuqi2020}. However, these mini-batching MCMC samplers usually take long mixing steps and are hard to calibrate, which contradicts the gist of using MCMC kernels in SMC: An ideal SMC sampler should need as few mixing steps as possible. Furthermore, the variance of the estimator increases in $n$, and it is likely that we need to adaptively increase the batch size to control the error. However, it turns out that we can rectify this problem by moving the stochastic mini-batching approximation outside of the Markov kernel to that of the gradient step as explained in the next section.
\looseness=-1

\subsection{Stochastic gradient sequential Monte Carlo}
\label{sec:sg-smc}
To make Algorithm~\ref{alg:standard-smc} scalable in the number of data points $N$, it is natural to approximate the marginal log-likelihood by a stochastic approximation, as in stochastic gradient descent. Let $1 \leq M \leq N$ denote the batch size and let $S_M\coloneqq \lbrace S_M(1), S_M(2), \ldots, S_M(M)\rbrace$ be a sequence of independent random integers (uniformly distributed in $[1, N]$) that represent the batch indices. We may approximate the marginal log-likelihood by $\log p(y_{1:N}; \psi) \approx N / \, M \log p(y_{S_M}; \psi)$, where $y_{S_M} \coloneqq \lbrace y_{S_M(1)}, y_{S_M(2)}, \ldots, y_{S_M(M)} \rbrace$ represents the corresponding subdataset. The same also goes for the gradient computation. More specifically, 
\begin{equation}
	\begin{split}
		&\grad\log p(y_{1:N}; \psi) \\
		&\approx \frac{N}{M}\expecbig{\grad\log p(y_{S_M}; \psi)} \\
		&= \frac{N}{M}\int \expecbig{\grad\log p(\phi, y_{S_M}; \psi) \, p(\phi \cond y_{S_M}; \psi)} \diff \phi,
	\end{split}
	\label{equ:sgsmc-gradient}
\end{equation}
where the expectation is taken over $S_M$. However, due to the latent variable $\phi$, the approximation $N / \, M \log p(y_{S_M}; \psi)$ is biased. Consequently, the gradient is also biased, and the bias scales in the difference between $p(\phi \cond y_{S_M})$ and $p(\phi \cond y_{1:N})$. 

Using the approximation in Equation~\eqref{equ:sgsmc-gradient} amounts to running Algorithm~\ref{alg:standard-smc} on the subdataset $y_{S_M}$ and then using the posterior samples of $p(\phi \cond y_{S_M}; \psi)$ to compute the gradient $ \int \grad\log(\phi, y_{S_M}; \psi) \, p(\phi \cond y_{S_M}; \psi) \diff \phi$. We then arrive at the stochastic gradient SMC sampler summarised in Algorithm~\ref{alg:sgd-smc}. 

\begin{algorithm2e}[h]
	\SetAlgoLined
	\DontPrintSemicolon
	\KwInputs{The same as in Algorithm~\ref{alg:standard-smc}, and batch size $M$}
	\KwOutputs{The MLE estimate $\psi_i$}
	\For{$i=1,2,\ldots$ \KwUntil convergent}{%
		Draw $\lbrace \phi_{0, j} \rbrace_{j=1}^J \sim \pi(\phi)$\;
		$w_{0, j} = 1 \, / \, J$ for all $j=1,2,\ldots,J$\;
		Draw subdataset $y_{S_M^i} \subseteq y_{1:N}$\;
		\For(\tcp*[f]{Parallelise $j$}){$n=1$ \KwTo $M$}{%
			Resample $\lbrace (w_{n-1, j}, \phi_{n-1, j})\rbrace_{j=1}^J$ if needed\;
			Draw $\phi_{n, j} \cond \phi_{n-1, j} \sim h_n(\phi_{n, j} \cond \phi_{n, j-1})$\;
			$\overline{w}_{n, j} = w_{n-1, j} \, p(y_{S_M^i(n)} \cond \phi_{n, j}; \psi_{i-1})$\;
			$w_{n, j} = \overline{w}_{n, j} \, / \, \sum_{k=1}^J \overline{w}_{n, k}$\;
		}
		$g(\psi_{i-1}) = \frac{N}{M}\sum_{j=1}^J w_{M, j}  \grad\log p(y_{S_M^i} \cond \phi_{M, j}; \psi)$\;
		$\psi_i = \psi_{i-1} + r(i) \, g(\psi_{i-1})$
	}
	\caption{Stochastic gradient sequential Monte Carlo (SGSMC) sampler for pBNN}
	\label{alg:sgd-smc}
\end{algorithm2e}

Compared to Algorithm~\ref{alg:standard-smc}, the stochastic gradient version in Algorithm~\ref{alg:sgd-smc} does not have to load the entire dataset to compute the gradient. More importantly, this also handles the Markov kernel design dilemma in Section~\ref{sec:markov-kernel} in the following way. The kernel $h_n$ is now chosen to be invariant with respect to $p(\phi \cond y_{S_M(1)}, \ldots, y_{S_M(n-1)})$ which is far easier to compute than the original when $M \ll N$. Essentially, the algorithm is a direct application of the stochastic gradient method on a latent variable model by using SMC samplers to approximate the gradient. This optimisation scheme is akin to the variational SMC method~\citep{Naesseth2018VSMC}.

\subsection{Open-horizon sequential Monte Carlo}
\label{sec:ohsmc}
Algorithm~\ref{alg:sgd-smc} defines an approximate flow of gradient that can be computed efficiently by applying SMC samplers in a closed data horizon. However, the algorithm does not directly output the target posterior distribution $p(\phi \cond y_{1:N}; \psi)$ unlike Algorithm~\ref{alg:standard-smc}. To compute the posterior distribution, we may need to run another Bayesian inference (e.g., by SMC or MCMC) based upon the estimated $\psi$. Moreover, for every optimisation step, the SMC estimators are independent and cold-start from the prior $\pi$. This causes a waste of computation, since Algorithm~\ref{alg:sgd-smc} does compute the posterior distributions $p(\phi \cond y_{S_M}; \psi_i)$ at subdatasets which are approximations to the target posterior distribution. In light of this observation, we can make Algorithm~\ref{alg:sgd-smc} even more efficient by linking the posterior distribution estimates in conjunction with the gradient. More specifically, we modify Algorithm~\ref{alg:sgd-smc} by warm-starting each SMC sampler from the previous posterior distribution estimate and perform the gradient update in conjunction with the SMC sampler. We then arrive at an SMC sampler that simultaneously estimates the posterior distribution and the parameter, summarised in Algorithm~\ref{alg:open-horizon-smc}. 

\begin{algorithm2e}[h]
	\SetAlgoLined
	\DontPrintSemicolon
	\KwInputs{Same as in Algorithm~\ref{alg:sgd-smc}}
	\KwOutputs{Same as in Algorithm~\ref{alg:standard-smc}}
	Draw $\lbrace \phi_{0, j} \rbrace_{j=1}^J \sim \pi(\phi)$\;
	$w_{0, j} = 1 \, / \, J$ for all $j=1,2,\ldots,J$\;
	\For(\tcp*[f]{Parallelise $j$}){$i=1,2,\ldots$ \KwUntil convergent}{%
		Draw subdataset $y_{S_M^i} \subseteq y_{1:N}$\;
		Resample $\lbrace (w_{i-1, j}, \phi_{i-1, j})\rbrace_{j=1}^J$ if needed\;
		Draw $\phi_{i, j} \cond \phi_{i-1, j} \sim h_i(\phi_{i, j} \cond \phi_{i, j-1})$\;
		$\overline{w}_{i, j} = w_{i-1, j} \, p(y_{S_M^i} \cond \phi_{i, j}; \psi_{i-1})$\;
		$w_{i, j} = \overline{w}_{i, j} \, / \, \sum_{s=1}^J \overline{w}_{i, s}$\;
		$g(\psi_{i-1}) = \frac{N}{M}\sum_{j=1}^J w_{i,j}\! \grad\log p(y_{S_M^i} \cond \phi_{i, j}; \psi_{i-1})$\;
		$\psi_i = \psi_{i-1} + r(i) \, g(\psi_{i-1})$
	}
	\caption{Open-horizon sequential Monte Carlo (OHSMC) sampler}
	\label{alg:open-horizon-smc}
\end{algorithm2e}

In Algorithm~\ref{alg:open-horizon-smc}, we start from samples drawn from the given prior $\pi$. Then, at each iteration $i$, we randomly draw a subdataset $y_{S_M^i}$ of $y_{1:N}$ and compute the posterior distribution and gradient on the subdataset by using the previous estimate. Equivalently, the algorithm is an SMC sampler applied on a growing dataset with open horizon instead of $y_{1:N}$ which has a fixed size~\citep[cf.][Eq. 5.6]{Kantas2015}. Compared to Algorithms~\ref{alg:standard-smc} and~\ref{alg:sgd-smc}, this open-horizon SMC sampler is computationally more efficient, especially when $N$ is large.\looseness=-1 

However, Algorithm~\ref{alg:open-horizon-smc} no longer targets the posterior distribution $p(\phi \cond y_{1:N}; \psi)$. To see what the algorithm does, let us \emph{fix} the parameter $\psi$, to find that the algorithm computes the following Feynman--Kac model
\begin{equation}
	\begin{split}
		&\frac{1}{\hat{l}_P(\psi)}\int \prod_{i=1}^P p(y_{S_M^i} \cond \phi_i; \psi) \prod _{i=1}^P h_i(\phi_i \cond \phi_{i-1}) \\
		&\qquad\qquad\quad\times \pi(\phi_0) \diff \phi_{0:P-1}, 
	\end{split}
	\label{equ:ohsmc-feynman-kac}
\end{equation}
where $P$ is the number of iterations for the algorithm, and $\hat{l}_P$ is the normalising constant. Evidently, Equation~\eqref{equ:ohsmc-feynman-kac} is neither equal to the original Feynman--Kac model in Equation~\eqref{equ:feynman-kac}, nor to its expectation. It is, however, possible to use Poisson estimators~\citep{Beskos2006, Fearnhead2008, Pierre2015} to make Equation~\eqref{equ:ohsmc-feynman-kac} an unbiased estimator of the original Feynman--Kac model. To do so, we need to let the number of iterations $P\sim \mathrm{Poisson}(\lambda)$ be a Poisson random variable and modify the weight update in Algorithm~\ref{alg:open-horizon-smc} to
\begin{equation*}
	\overline{w}_{i, j} = w_{i-1, j} \, \Bigl(\frac{N}{M}\log p(y_{S_M^i} \cond \phi_{i, j}; \psi_{i-1}) + \log c\Bigr),
\end{equation*}
where $c$ is a constant that guarantees the positivity of the weights almost surely. Then, we can show by the Poisson estimator that 
\begin{equation*}
	\begin{split}
		&\expecbigg{\frac{\expp^{\lambda}}{c}\prod_{i=1}^P \frac{1}{\lambda} \, \Bigl(\frac{N}{M}\log p(y_{S_M^i} \cond \phi; \psi) + \log c\Bigr)} \\
		&= \prod^{N}_{i=1} p(y_{i} \cond \phi; \psi),
	\end{split}
\end{equation*}
where the expectation is taken on both $P$ and $S_M$. However, finding such a positive constant $c$ is still an open but stimulating challenge in the statistic community~\cite[see, e.g., a recent progress in][]{jin2022biasing}. 

For training pBNNs, it may not be necessary to debias Algorithm~\ref{alg:open-horizon-smc}. 
In practice, we use (p)BNNs as powerful predictive models to quantify uncertainty. 
Hence, computing the exact posterior distribution is often unnecessary from a practical point of view~\citep{Wilson2020, izmailov2021bayesian}, and we can control the bias via validation data. 
As a consequence, the prior is no longer considered as given but is a flexible model free to choose so as to aim for better predictive performance. 
To this end, we can further relax the invariance property of the Markov kernel $h$. This relaxation bridges a connection to the classical stochastic filtering approaches for optimisations~\citep{Bell1994, Gerber2021}. 
As an example, the extended Kalman filter approximations to the algorithm are stochastic natural gradient descents with information matrices determined by the Markov kernel~\citep{Ollivier2018, Martens2020}. 
As another example, if we choose the Markov kernel to be that of a Brownian motion, then the resulting algorithm is akin to using stochastic filters for training neural networks with uncertainties~\citep{Singhal1988, Freitas2000, Chang2022}. 
It is also possible to debias the Brownian motion kernel by that of a backward kernel as in~\citet{Maqrues2013} and~\citet{DelMoral2006}. 

It is worth remarking a minor practical advantage of Algorithm~\ref{alg:open-horizon-smc} over Algorithm~\ref{alg:sgd-smc}. The Markov kernel in Algorithm~\ref{alg:sgd-smc} is a function that has dynamic input sizes. This means that implementing the algorithm in commonly used automatic differentiation libraries that assume static input shapes (e.g., JAX and Tensorflow) is not trivial. On the other hand, Algorithm~\ref{alg:open-horizon-smc} has no such an issue. 

\section{EXPERIMENTS}
\label{sec:experiments}
In this section, we evaluate our proposed Algorithms~\ref{alg:sgd-smc} and~\ref{alg:open-horizon-smc} which we call SGSMC and OHSMC, respectively, in several ways. First, we test the methods on a synthetic model to see if they can recover the true parameters and posterior distribution. 
Then, we train pBNNs using the methods for regression and classification tasks on synthetic, UCI, and MNIST datasets. Our implementation is publicly available at \url{https://github.com/spdes/pbnn}.
\paragraph{Baselines} We compare against (i) the maximum a posteriori (MAP) method for estimating the parameter and use Hamiltonian Monte Carlo (MAP-HMC) to compute the posterior distribution based on the learnt parameter~\citep[][Sec. 6]{Sharma2023}, (ii) stochastic weight averaging Gaussian~\citep[SWAG,][]{maddox2019simple} with the MAP objective function, and (iii) stochastic mean-field Gaussian variational Bayes~\citep[VB,][]{Hoffman2013}. In addition, SGSMC-HMC refers to sampling the posterior distribution by HMC and estimating the parameters by SGSMC. \looseness=-1
\paragraph{Setting} We use Adam for all methods with a learning rate of 0.01 unless stated otherwise. As for the prior, we consistently employ a standard Gaussian distribution. Whenever sampling from the posterior distribution is needed, we use $J=1,000$ samples. We use the same amount for evaluations. For SGSMC, we use an MCMC random walk kernel, whereas for OHSMC, we use a random walk kernel (with variance 0.01). VB uses 100 MC samples to approximate the evidence lower bound. Further details (e.g., batch sizes, number of epochs, and pBNN structures) for all the experiments are found in the appendix. \looseness=-1

\begin{figure*}
	\begin{minipage}{.32\linewidth}
		\includegraphics[width=\linewidth]{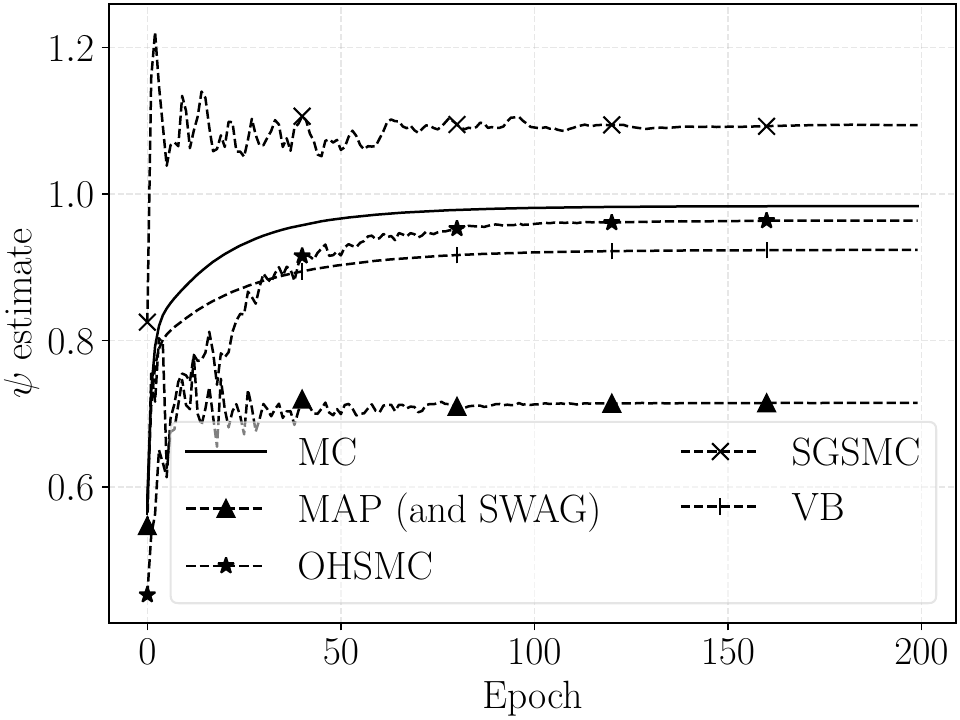}
		\caption{Traces of parameter estimations. The true value is 1.}
		\label{fig:crescent-param-est}
	\end{minipage}
	\hfill
	\begin{minipage}{.32\linewidth}
		\includegraphics[width=\linewidth]{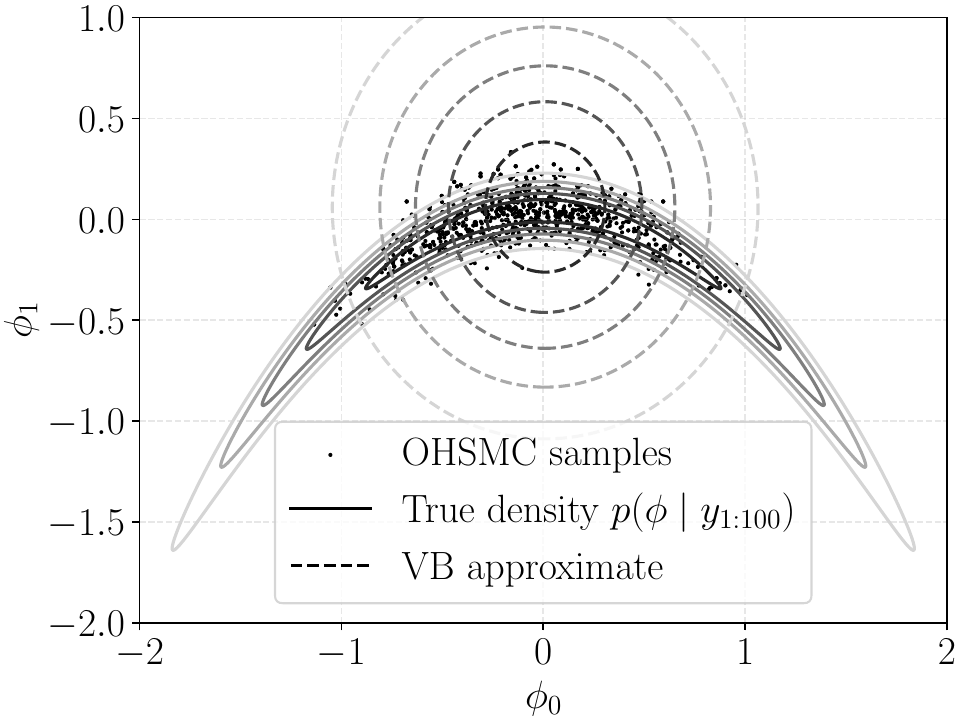}
		\caption{Posterior approximation for the model in Equation~\eqref{equ:crescent}.}
		\label{fig:crescent-density}
	\end{minipage}
	\hfill
	\begin{minipage}{.32\linewidth}
		\renewcommand{\figurename}{Table}
		\setcounter{figure}{0} 
		\caption{One-dimensional regression results. The best result for each column is bold.}
		\label{tbl:reg-errs}
		\setcounter{table}{1} 
		\setcounter{figure}{2} 
		\renewcommand{\figurename}{Figure}
		\center
		\resizebox{\linewidth}{!}{%
			\begin{tabular}{@{}lll@{}}
				\toprule
				Method  & NLPD (std.)       & RMSE (std.)   \\ \midrule
				MAP-HMC & 1.53 (0.09) & 0.59 (0.11) \\
				OHSMC   & \textbf{1.49} (0.07) & \textbf{0.53} (0.07) \\
				SGSMC-HMC   & 1.65 (0.16) & 0.75 (0.14) \\
				SWAG    & 1.71 (0.14) & 0.82 (0.16) \\
				VB      & 2.13 (0.15) & 1.46 (0.46) \\ \bottomrule
			\end{tabular}
		}
	\end{minipage}
\end{figure*}

\subsection{Synthetic parameter estimation}
\label{sec:syn-param-est}
Consider a model given by
\begin{equation}
	\begin{split}
		\begin{bmatrix}
			\phi_0 \\ 
			\phi_1
		\end{bmatrix}
		&\sim 
		\mathrm{N}\biggl(
		\begin{bmatrix}
			0\\0
		\end{bmatrix}, 
		\begin{bmatrix}
			2 & 0\\
			0 & 1
		\end{bmatrix}
		\biggr), \\
		y_n \cond \phi &\sim \mathrm{N}\Bigl(\frac{\phi_1}{\psi} + \frac{1}{2} \, (\phi_0^2 + \psi^2), 1 \Bigr), 
		\label{equ:crescent}
	\end{split}
\end{equation}
where we generate a sequence of 100 independent data points $y_{1:100}$ under the true parameter $\psi = 1$ and the realisation $\phi = \begin{bmatrix} 0 & 0 \end{bmatrix}$. The goal is to estimate the parameter $\psi$ and also to compute the posterior distribution which has a non-Gaussian crescent shape. For this model, we can conveniently compute a tight lower bound of the MLE objective by brute-force Monte Carlo with 10,000 samples, which we call MC. However, for training pBNNs, the MC approach is in general not applicable.

In Figure~\ref{fig:crescent-param-est}, we see that the MC estimate is closest to the truth followed by OHSMC, VB, and SGSMC. The MAP estimator on the other hand, significantly diverges from the truth. Moreover, from Figure~\ref{fig:crescent-density} we see that the OHSMC samples are close to the true density, although the samples do not explore the tails of the true distribution. The VB estimate is correct in the mean approximation, but the Gaussian approximation does not fit the shape of the true distribution. The SWAG method, which uses the results based on the MAP estimate, produces a covariance matrix whose diagonal is numerically zero.

\subsection{Synthetic regression and classification}
\label{sec:syn-reg-cla}
\paragraph{Regression}
Next, we benchmark the algorithms on a synthetic regression problem where we have access to the true underlying function. The training, validation, and test data are generated as per $y_n = f(x_n) + \xi_n$, for $n=1,2,\ldots, 100$, where $f(x) = x \sin(x \tanh(x))$, and i.i.d. noise $\xi_n\sim \mathrm{N}(0, 1)$. We use a three-layer pBNN (with output sizes 20, 10, and 1), where the stochastic part is on the second hidden layer. We repeat the experiment 100 times and report the averaged negative log predictive density (NLPD) on the test data, and root mean-square error (RMSE) on the true function including their standard deviations.

\begin{figure*}[t!]
	\centering
	\includegraphics[width=.99\linewidth]{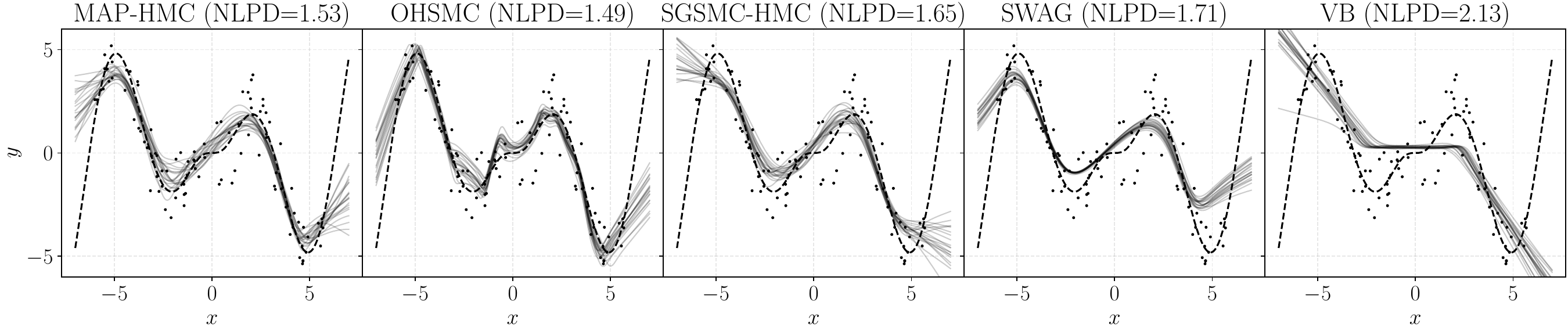}
	\caption{Visualisation of the synthetic regression. The scatter points represent the test data and the dashed line depicts the true function. The grey lines are predictive samples from their learnt pBNNs.}
	\label{fig:reg-vis}
\end{figure*}

Table~\ref{tbl:reg-errs} shows that the proposed OHSMC method performs best in both metrics. Furthermore, Figure~\ref{fig:reg-vis} shows that the predictive samples drawn from the pBNNs learnt by OHSMC best capture the local optima of the true function. Moreover, OHSMC extrapolates the regression problem best.

\paragraph{Classification}
As for classification, we test the methods on a synthetic two-moon dataset. Apart from NLPD, we also report the expected calibration error~\citep[ECE,][]{GuoECE2017} and accuracy. The utilised neural network has four dense layers (with output sizes 100, 20, 5, and 1), where the stochastic part is the third layer. 
\looseness=-1

\begin{table}[t!]
	\caption{Classification results in two-moon data.}
	\label{tbl:moon}
	\center
	\resizebox{.99\linewidth}{!}{%
	\begin{tabular}{@{}llll@{}}
		\toprule
		Method    & NLPD (std.) & ECE (std.)  & Acc. (std.) \\ \midrule
		MAP-HMC   & \textbf{0.28} (0.06) & 0.07 (0.01) & 0.87 (0.02)     \\
		OHSMC     & \textbf{0.28} (0.07) & \textbf{0.06} (0.01) & \textbf{0.88} (0.02)     \\
		SGSMC-HMC & 0.32 (0.08) & 0.08 (0.01) & 0.86 (0.03)     \\
		SWAG      & 0.31 (0.06) & 0.07 (0.01) & 0.86 (0.03)     \\
		VB        & 0.29 (0.05) & 0.08 (0.01) & 0.86 (0.03)     \\ \bottomrule
		\end{tabular}
	}
\end{table}

Table~\ref{tbl:moon} shows that our OHSMC either performs better than its peers or on par in terms of NLPD. According to the ECE and accuracy metrics, OHSMC is best.
\looseness=-1

\subsection{UCI regression and classification}
\label{sec:uci}
We now move from synthetic experiments to real-world UCI data~\citep{uci}.
Throughout, we use a neural network with four dense layers (with output sizes 50, 20, 5, and $C$, where $C$ is the number of labels), and place the stochastic part on the third layer. All experiments are repeated ten times and we report averaged metrics. 

Table~\ref{tbl:uci} shows the results on two regression and two classification UCI datasets. As before, our proposed OHSMC method either outperforms or performs on par with other methods in terms of NLPD for all datasets. As for regression, the RMSE is significantly lower for OHSMC. Moreover, our methods are marginally better than the others when evaluating ECE for the classification tasks. Results on additional UCI datasets are found in Appendix~\ref{appendix:uci}.

\begin{figure*}
	\centering
	\includegraphics[width=.99\linewidth]{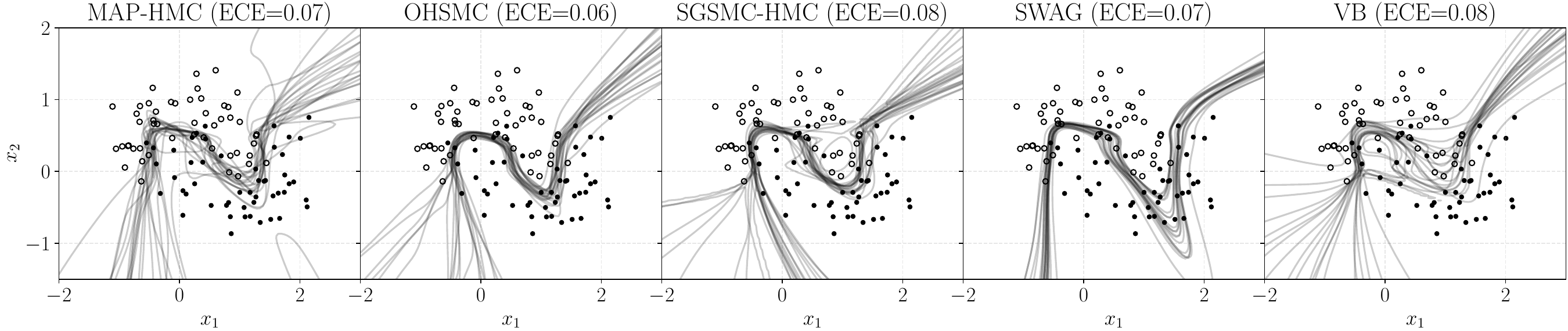}
	\caption{Visualisation of the two-moons classifications. The scatter points are the test data with hollow/solid representing the label. The grey lines represent the classification hyperplanes sampled from the trained pBNNs. }
	\label{fig:two-moons}
\end{figure*}

\begin{table*}[t!]
	\caption{Results on two regression and two classification UCI datasets. Results on more UCI datasets are provided in Table~\ref{tbl:uci-data} in Appendix~\ref{appendix:uci}. }
	\label{tbl:uci}
	\begin{center}
		\resizebox{.99\linewidth}{!}{%
		\begin{tabular}{rrrrrrrrr}
			\toprule
			\multirow{2}{*}{Method} & \multicolumn{2}{c}{yacht} & \multicolumn{2}{c}{energy} & \multicolumn{2}{c}{satellite} & \multicolumn{2}{c}{ionosphere} \\ \cmidrule(l){2-3} \cmidrule(l){4-5} \cmidrule(l){6-7} \cmidrule(l){8-9}  
			               & NLPD                 & RMSE                  & NLPD         & RMSE              & NLPD        & ECE          & NLPD           & ECE           \\ \midrule
MAP-HMC        & 0.95 (0.01)          & 0.27 (0.04)           & 0.94 (0.00)           & 0.24 (0.02)                & 0.28 (0.03) & 0.04 (0.00)  & \textbf{0.21} (0.09)    & 0.09 (0.03)   \\
OHSMC          & \textbf{0.92} (0.00) & \textbf{0.11} (0.04)  & \textbf{0.92} (0.00)  & \textbf{0.09} (0.00)       & \textbf{0.27} (0.03) & \textbf{0.03} (0.00)  & \textbf{0.21} (0.13)    & \textbf{0.08} (0.03)   \\
SGSMC-HMC          & 0.93 (0.00)          & 0.17 (0.02)           & \textbf{0.92} (0.00)  & 0.14 (0.02)                & 0.84 (0.70) & 0.10 (0.10)  & 0.24 (0.19)    & \textbf{0.08} (0.04)   \\
SWAG           & 0.95 (0.02)          & 0.25 (0.08)           & 0.95 (0.00)           & 0.25 (0.02)                & 0.35 (0.07) & 0.07 (0.03)  & 0.36 (0.12)    & 0.14 (0.05)   \\
VB             & 0.94 (0.00)          & 0.20 (0.02)           & 0.93 (0.00)           & 0.19 (0.03)                & 0.28 (0.04) & \textbf{0.03} (0.00)  & 0.28 (0.16)    & 0.10 (0.05)   \\ \bottomrule
		\end{tabular}
		}
	\end{center}
\end{table*}

\subsection{MNIST classification}
\label{sec:mnist}
We also consider classification on MNIST data~\citep{lecun1998mnist} 
by using a neural network with two convolutional layers followed by two dense layers. The stochastic part is on the first convolution layer. Due to their high computational cost, HMC methods do not apply here. Table~\ref{tbl:mnist} shows that our OHSMC method significantly outperforms other methods according to all the three metrics. 

\begin{table}[t!]
	\caption{Classification results on MNIST. Within the table, NLPD and ECE are scaled by $10^{-2}$. }
	\label{tbl:mnist}
	\begin{center}
		\resizebox{.99\linewidth}{!}{%
			\begin{tabular}{@{}llll@{}}
				\toprule
				Method & NLPD (std.) & ECE (std.) & Acc. (std.) \\ \midrule
				OHSMC  &  \textbf{2.87} (0.24)  &    \textbf{0.35} (0.06)        &    \textbf{99.02\%} (0.07)             \\
				SWAG   &  4.27 (0.83)           &    1.17 (0.43)        &    97.76\% (0.71)             \\
				VB     &  4.53 (0.28)           &    0.44 (0.08)        &   98.51\% (0.09)              \\ \bottomrule
			\end{tabular}
		}
	\end{center}
\end{table}

In this MNIST task, OHSMC (with 100 samples), SWAG, and VB (with 100 samples) take around 0.80, 0.26, and 0.33 seconds, respectively, for running 1,000 iterations with batch size 20. Note that the time for OHSMC includes \emph{both}, the parameter estimation and posterior sampling, while for SWAG the time is only for parameter estimation. The times are profiled on an NVIDIA A100 40GB GPU. 

\section{CONCLUSIONS}
\label{sec:conclusion}
In this paper, we have shown a sequential Monte Carlo (SMC) routine to efficiently train partial Bayesian neural networks (pBNNs). Specifically, we have proposed two approximate SMC samplers that are suitable for estimating the parameters and posterior distributions of pBNNs. The proposed training scheme either outperforms, or performs on par with state-of-the-art methods on a variety of datasets. 

\paragraph{Limitations and future work}
%The proposed training scheme is immediately parallelisable on GPUs but is potentially at the cost of a high memory usage.
%An interesting future work is to theoretically analyse the convergence of Algorithm~\ref{alg:open-horizon-smc} as a general method for inference in latent variable models, since the algorithm carries a joint flow of both gradient and distribution. 
Algorithm~\ref{alg:open-horizon-smc} does not necessarily converge to the target distribution (under \emph{specified} prior and likelihood models) due to its bias. However, the bias in the pBNN context is not an issue when we opt for the predictive performance as we have empirically demonstrated in our experiments. As future work, it is interesting to study how the algorithm performs for inference in more general latent variable models, and to analyse its theoretical convergence properties.

\subsubsection*{Acknowledgments}
The authors would like to thank Adrien Corenflos for his comments. This work was partially supported by the Wallenberg AI, Autonomous Systems and Software Program (WASP) funded by the Knut and Alice Wallenberg Foundation, and by the Kjell och M\"{a}rta Beijer Foundation. The computations/data handling were enabled by the supercomputing resource Berzelius provided by National Supercomputer Centre at Link\"{o}ping University and the Knut and Alice Wallenberg foundation. \looseness=-1

\bibliography{refs}

\begin{thebibliography}{58}
\providecommand{\natexlab}[1]{#1}
\providecommand{\url}[1]{\texttt{#1}}
\expandafter\ifx\csname urlstyle\endcsname\relax
  \providecommand{\doi}[1]{doi: #1}\else
  \providecommand{\doi}{doi: \begingroup \urlstyle{rm}\Url}\fi

\bibitem[Andrieu and Roberts(2009)]{Andrieu2009}
Christophe Andrieu and Gareth~O. Roberts.
\newblock The pseudo-marginal approach for efficient {M}onte {C}arlo
  computations.
\newblock \emph{The Annals of Statistics}, 37\penalty0 (2):\penalty0 697--725,
  2009.

\bibitem[Arya et~al.(2022)Arya, Schauer, Sch\"{a}fer, and
  Rackauckas]{Gaurav2022}
Gaurav Arya, Moritz Schauer, Frank Sch\"{a}fer, and Christopher Rackauckas.
\newblock Automatic differentiation of programs with discrete randomness.
\newblock In \emph{Proceedings of Advances in Neural Information Processing
  Systems}, volume~35, pages 10435--10447. Curran Associates, Inc., 2022.

\bibitem[Arya et~al.(2023)Arya, Seyer, Sch\"{a}fer, Chandra, Lew, Huot,
  Mansinghka, Ragan-Kelley, Rackauckas, and Schauer]{Gaurav2023}
Gaurav Arya, Ruben Seyer, Frank Sch\"{a}fer, Kartik Chandra, Alexander~K. Lew,
  Mathieu Huot, Vikash~K. Mansinghka, Jonathan Ragan-Kelley, Christopher
  Rackauckas, and Moritz Schauer.
\newblock Differentiating {M}etropolis-{H}astings to optimize intractable
  densities.
\newblock In \emph{The 40th International Conference on Machine Learning
  Workshop: Differentiable Almost Everything}, 2023.

\bibitem[Bell(1994)]{Bell1994}
Bradley~M. Bell.
\newblock The iterated {K}alman smoother as a {G}auss--{N}ewton method.
\newblock \emph{SIAM Journal on Optimization}, 4\penalty0 (3):\penalty0
  626--636, 1994.

\bibitem[Beskos et~al.(2006)Beskos, Papaspiliopoulos, and Roberts]{Beskos2006}
Alexandros Beskos, Omiros Papaspiliopoulos, and Gareth~O. Roberts.
\newblock Retrospective exact simulation of diffusion sample paths with
  applications.
\newblock \emph{Bernoulli}, 12\penalty0 (6):\penalty0 1077--1098, 2006.

\bibitem[Bishop(2006)]{Bishop2006Book}
Christopher~M. Bishop.
\newblock \emph{Pattern recognition and machine learning}.
\newblock Springer, 2006.

\bibitem[Blei et~al.(2017)Blei, Kucukelbir, and McAuliffe]{blei2017variational}
David~M. Blei, Alp Kucukelbir, and Jon~D. McAuliffe.
\newblock Variational inference: A review for statisticians.
\newblock \emph{Journal of the American Statistical Association}, 112\penalty0
  (518):\penalty0 859--877, 2017.

\bibitem[Blundell et~al.(2015)Blundell, Cornebise, Kavukcuoglu, and
  Wierstra]{Blundell2015}
Charles Blundell, Julien Cornebise, Koray Kavukcuoglu, and Daan Wierstra.
\newblock Weight uncertainty in neural network.
\newblock In \emph{Proceedings of the 32nd International Conference on Machine
  Learning}, volume~37, pages 1613--1622. PMLR, 2015.

\bibitem[Bradbury et~al.(2018)Bradbury, Frostig, Hawkins, Johnson, Leary,
  Maclaurin, Necula, Paszke, VanderPlas, Wanderman-Milne, and
  Zhang]{jax2018github}
James Bradbury, Roy Frostig, Peter Hawkins, Matthew~James Johnson, Chris Leary,
  Dougal Maclaurin, George Necula, Adam Paszke, Jake VanderPlas, Skye
  Wanderman-Milne, and Qiao Zhang.
\newblock {JAX}: composable transformations of {P}ython+{N}um{P}y programs,
  2018.
\newblock URL \url{http://github.com/google/jax}.

\bibitem[Cabezas et~al.(2023)Cabezas, Lao, and Louf]{blackjax2020github}
Alberto Cabezas, Junpeng Lao, and R\'emi Louf.
\newblock {B}lackjax: a sampling library for {JAX}, 2023.
\newblock URL \url{http://github.com/blackjax-devs/blackjax}.

\bibitem[Capp\'{e} et~al.(2005)Capp\'{e}, Moulines, and Ryd\'{e}n]{Cappe2005}
Olivier Capp\'{e}, Eric Moulines, and Tobias Ryd\'{e}n.
\newblock \emph{Inference in hidden {M}arkov models}.
\newblock Springer Series in Statistics. Springer-Verlag, 2005.

\bibitem[Chang et~al.(2022)Chang, Jone, and Murphy]{Chang2022}
Peter~G. Chang, Matt Jone, and Keven Murphy.
\newblock On diagonal approximations to the extended {K}alman filter for online
  training of {B}ayesian neural networks.
\newblock In \emph{Proceedings of the 14th Asian Conference on Machine Learning
  Workshop}. OpenReview, 2022.

\bibitem[Chen et~al.(2014)Chen, Fox, and Guestrin]{ChenTQ2014}
Tianqi Chen, Emily Fox, and Carlos Guestrin.
\newblock Stochastic gradient {H}amiltonian {M}onte {C}arlo.
\newblock In \emph{Proceedings of the 31st International Conference on Machine
  Learning}, volume~32, pages 1683--1691. PMLR, 2014.

\bibitem[Chopin(2002)]{Chopin2002}
Nicolas Chopin.
\newblock A sequential particle filter method for static models.
\newblock \emph{Biometrika}, 89\penalty0 (3):\penalty0 539--551, 2002.

\bibitem[Chopin and Papaspiliopoulos(2020)]{ChopinBook2020}
Nicolas Chopin and Omiros Papaspiliopoulos.
\newblock \emph{An introduction to sequential {M}onte {C}arlo}.
\newblock Springer Series in Statistics. Springer Nature Switzerland, 2020.

\bibitem[Corenflos et~al.(2021)Corenflos, Thornton, Deligiannidis, and
  Doucet]{Corenflos2011}
Adrien Corenflos, James Thornton, George Deligiannidis, and Arnaud Doucet.
\newblock Differentiable particle filtering via entropy-regularized optimal
  transport.
\newblock In \emph{Proceedings of the 38th International Conference on Machine
  Learning}, volume 139, pages 2100--2111. PMLR, 2021.

\bibitem[Dau and Chopin(2021)]{Dau2021}
Hai-Dang Dau and Nicolas Chopin.
\newblock Waste-free sequential {M}onte {C}arlo.
\newblock \emph{Journal of the Royal Statistical Society Series B: Statistical
  Methodology}, 84\penalty0 (1):\penalty0 114--148, 2021.

\bibitem[Daxberger et~al.(2021)Daxberger, Nalisnick, Allingham, Antor{\'a}n,
  and Hern{\'a}ndez-Lobato]{daxberger2021bayesian}
Erik Daxberger, Eric Nalisnick, James~U. Allingham, Javier Antor{\'a}n, and
  Jos{\'e}~Miguel Hern{\'a}ndez-Lobato.
\newblock Bayesian deep learning via subnetwork inference.
\newblock In \emph{Proceedings of the 38th International Conference on Machine
  Learning}, volume 139, pages 2510--2521. PMLR, 2021.

\bibitem[De~Freitas et~al.(2000)De~Freitas, Niranjan, Gee, and
  Doucet]{Freitas2000}
Jo{\~a}o De~Freitas, Mahesan Niranjan, Andrew~H. Gee, and Arnaud Doucet.
\newblock Sequential {M}onte {C}arlo methods to train neural network models.
\newblock \emph{Neural Computation}, 12\penalty0 (4):\penalty0 955--993, 2000.

\bibitem[Del~Moral et~al.(2006)Del~Moral, Doucet, and Jasra]{DelMoral2006}
Pierre Del~Moral, Arnaud Doucet, and Ajay Jasra.
\newblock Sequential {M}onte {C}arlo samplers.
\newblock \emph{Journal of the Royal Statistical Society Series B: Statistical
  Methodology}, 68\penalty0 (3):\penalty0 411--436, 2006.

\bibitem[Fearnhead et~al.(2008)Fearnhead, Papaspiliopoulos, and
  Roberts]{Fearnhead2008}
Paul Fearnhead, Omiros Papaspiliopoulos, and Gareth~O. Roberts.
\newblock Particle filters for partially observed diffusions.
\newblock \emph{Journal of the Royal Statistical Society Series B: Statistical
  Methodology}, 70\penalty0 (4):\penalty0 755--777, 2008.

\bibitem[Foong et~al.(2020)Foong, Burt, Li, and
  Turner]{foong2020expressiveness}
Andrew Foong, David Burt, Yingzhen Li, and Richard Turner.
\newblock On the expressiveness of approximate inference in {B}ayesian neural
  networks.
\newblock In \emph{Proceedings of Advances in Neural Information Processing
  Systems}, volume~33, pages 15897--15908, 2020.

\bibitem[Gal and Ghahramani(2016)]{Gal2016}
Yarin Gal and Zoubin Ghahramani.
\newblock Dropout as a {B}ayesian approximation: Representing model uncertainty
  in deep learning.
\newblock In \emph{Proceedings of The 33rd International Conference on Machine
  Learning}, volume~48, pages 1050--1059. PMLR, 2016.

\bibitem[Gerber and Douc(2021)]{Gerber2021}
Mathieu Gerber and Randal Douc.
\newblock A global stochastic optimization particle filter algorithm.
\newblock \emph{Biometrika}, 109\penalty0 (4):\penalty0 937--955, 2021.

\bibitem[Golub and Meurant(2010)]{Golub2010}
Gene~H. Golub and G\'{e}ard Meurant.
\newblock \emph{Matrices, moments and quadrature with applications}.
\newblock Princeton series in applied mathematics. Princeton University Press,
  2010.

\bibitem[Guo et~al.(2017)Guo, Pleiss, Sun, and Weinberger]{GuoECE2017}
Chuan Guo, Geoff Pleiss, Yu~Sun, and Kilian~Q. Weinberger.
\newblock On calibration of modern neural networks.
\newblock In \emph{Proceedings of the 34th International Conference on Machine
  Learning}, volume~70, pages 1321--1330. PMLR, 2017.

\bibitem[Heek et~al.(2023)Heek, Levskaya, Oliver, Ritter, Rondepierre, Steiner,
  and van Zee]{flax2020github}
Jonathan Heek, Anselm Levskaya, Avital Oliver, Marvin Ritter, Bertrand
  Rondepierre, Andreas Steiner, and Marc van Zee.
\newblock {F}lax: a neural network library and ecosystem for {JAX}, 2023.
\newblock URL \url{http://github.com/google/flax}.

\bibitem[Hoffman et~al.(2013)Hoffman, Blei, Wang, and Paisley]{Hoffman2013}
Matthew~D. Hoffman, David~M. Blei, Chong Wang, and John Paisley.
\newblock Stochastic variational inference.
\newblock \emph{Journal of Machine Learning Research}, 14\penalty0
  (40):\penalty0 1303--1347, 2013.

\bibitem[Izmailov et~al.(2018)Izmailov, Podoprikhin, Garipov, Vetrov, and
  Wilson]{izmailov2018averaging}
Pavel Izmailov, Dmitrii Podoprikhin, Timur Garipov, Dmitry Vetrov, and
  Andrew~Gordon Wilson.
\newblock Averaging weights leads to wider optima and better generalization.
\newblock In \emph{Proceedings of Conference on Uncertainty in Artificial
  Intelligence}, 2018.

\bibitem[Izmailov et~al.(2020)Izmailov, Maddox, Kirichenko, Garipov, Vetrov,
  and Wilson]{izmailov2020subspace}
Pavel Izmailov, Wesley~J. Maddox, Polina Kirichenko, Timur Garipov, Dmitry
  Vetrov, and Andrew~Gordon Wilson.
\newblock Subspace inference for {B}ayesian deep learning.
\newblock In \emph{Proceedings of the 35th Uncertainty in Artificial
  Intelligence Conference}, pages 1169--1179. PMLR, 2020.

\bibitem[Izmailov et~al.(2021)Izmailov, Vikram, Hoffman, and
  Wilson]{izmailov2021bayesian}
Pavel Izmailov, Sharad Vikram, Matthew~D. Hoffman, and Andrew~Gordon Wilson.
\newblock What are {B}ayesian neural network posteriors really like?
\newblock In \emph{Proceedings of the 38th International Conference on Machine
  Learning}, volume 139, pages 4629--4640. PMLR, 2021.

\bibitem[Jacob and Thiery(2015)]{Pierre2015}
Pierre~E. Jacob and Alexandre~H. Thiery.
\newblock On nonnegative unbiased estimators.
\newblock \emph{The Annals of Statistics}, 43\penalty0 (2):\penalty0 769--784,
  2015.

\bibitem[Jacob et~al.(2020)Jacob, O’Leary, and Atchad\'{e}]{Jacob2020}
Pierre~E. Jacob, John O’Leary, and Yves~F. Atchad\'{e}.
\newblock Unbiased {M}arkov chain {M}onte {C}arlo methods with couplings.
\newblock \emph{Journal of the Royal Statistical Society Series B: Statistical
  Methodology}, 82\penalty0 (3):\penalty0 543--600, 2020.

\bibitem[Jin et~al.(2022)Jin, Singh, and Chopin]{jin2022biasing}
Ruiyang Jin, Sumeetpal~S. Singh, and Nicolas Chopin.
\newblock De-biasing particle filtering for a continuous time hidden {M}arkov
  model with a {C}ox process observation model.
\newblock \emph{arXiv preprint arXiv:2206.10478}, 2022.

\bibitem[Johansen et~al.(2008)Johansen, Doucet, and Davy]{Johansen2008}
Adam~M. Johansen, Arnaud Doucet, and Manuel Davy.
\newblock Particle methods for maximum likelihood estimation in latent variable
  models.
\newblock \emph{Statistics and Computing}, 18\penalty0 (1):\penalty0 47--57,
  2008.

\bibitem[Kantas et~al.(2015)Kantas, Doucet, Singh, Maciejowski, and
  Chopin]{Kantas2015}
Nikolas Kantas, Arnaud Doucet, Sumeetpal~S. Singh, Jan Maciejowski, and Nicolas
  Chopin.
\newblock On particle methods for parameter estimation in state-space models.
\newblock \emph{Statistical Science}, 30\penalty0 (3):\penalty0 328--351, 2015.

\bibitem[Kelly et~al.(Accessed 2023)Kelly, Longjohn, and Nottingham]{uci}
Markelle Kelly, Rachel Longjohn, and Kolby Nottingham.
\newblock {UCI} machine learning repository, Accessed 2023.
\newblock URL \url{http://archive.ics.uci.edu/ml}.

\bibitem[Kristiadi et~al.(2020)Kristiadi, Hein, and Hennig]{kristiadi2020being}
Agustinus Kristiadi, Matthias Hein, and Philipp Hennig.
\newblock Being {B}ayesian, even just a bit, fixes overconfidence in {ReLU}
  networks.
\newblock In \emph{Proceedings of the 37th International Conference on Machine
  Learning}, volume 119, pages 5436--5446. PMLR, 2020.

\bibitem[LeCun et~al.(Accessed 2023)LeCun, Cortes, and Burges]{lecun1998mnist}
Yann LeCun, Corinna Cortes, and Christopher J.~C. Burges.
\newblock The {MNIST} database of handwritten digits, Accessed 2023.
\newblock URL \url{http://yann.lecun.com/exdb/mnist/}.

\bibitem[Lee et~al.(2010)Lee, Yau, Giles, Doucet, and Holmes]{Leee2010}
Anthony Lee, Christopher Yau, Michael~B. Giles, Arnaud Doucet, and
  Christopher~C. Holmes.
\newblock On the utility of graphics cards to perform massively parallel
  simulation of advanced {M}onte {C}arlo methods.
\newblock \emph{Journal of Computational and Graphical Statistics}, 19\penalty0
  (4):\penalty0 769--789, 2010.

\bibitem[Maddox et~al.(2019)Maddox, Izmailov, Garipov, Vetrov, and
  Wilson]{maddox2019simple}
Wesley~J. Maddox, Pavel Izmailov, Timur Garipov, Dmitry~P. Vetrov, and
  Andrew~Gordon Wilson.
\newblock A simple baseline for {B}ayesian uncertainty in deep learning.
\newblock In \emph{Proceedings of Advances in Neural Information Processing
  Systems}, volume~32, 2019.

\bibitem[Marques and Storvik(2013)]{Maqrues2013}
Reinaldo A.~Gomes Marques and Geir Storvik.
\newblock Particle move-reweighting strategies for online inference.
\newblock Technical report, University of Oslo and Statistics for Innovation
  Centre, 2013.

\bibitem[Martens(2020)]{Martens2020}
James Martens.
\newblock New insights and perspectives on the natural gradient method.
\newblock \emph{Journal of Machine Learning Research}, 21\penalty0
  (146):\penalty0 1--76, 2020.

\bibitem[Naesseth et~al.(2018)Naesseth, Linderman, Ranganath, and
  Blei]{Naesseth2018VSMC}
Christian Naesseth, Scott Linderman, Rajesh Ranganath, and David Blei.
\newblock Variational sequential {M}onte {C}arlo.
\newblock In \emph{Proceedings of the 21th International Conference on
  Artificial Intelligence and Statistics}, volume~84, pages 968--977. PMLR,
  2018.

\bibitem[Neal(2001)]{Neal2001}
Radford~M. Neal.
\newblock Annealed importance sampling.
\newblock \emph{Statistics and Computing}, 11\penalty0 (2):\penalty0 125--139,
  2001.

\bibitem[Ober and Rasmussen(2019)]{ober2019benchmarking}
Sebastian~W. Ober and Carl~Edward Rasmussen.
\newblock Benchmarking the neural linear model for regression.
\newblock In \emph{Proceedings of the 2nd Symposium on Advances in Approximate
  Bayesian Inference}, pages 1--25, 2019.

\bibitem[Ollivier(2018)]{Ollivier2018}
Yann Ollivier.
\newblock Online natural gradient as a {K}alman filter.
\newblock \emph{Electronic Journal of Statistics}, 12\penalty0 (2):\penalty0
  2930--2961, 2018.

\bibitem[Opper(1999)]{Opper1999}
Manfred Opper.
\newblock A {B}ayesian approach to on-line learning.
\newblock In \emph{On-line learning in neural networks}. Cambridge University
  Press, 1999.

\bibitem[Papamarkou et~al.(2022)Papamarkou, Hinkle, Young, and
  Womble]{Papamarkou2022}
Theodore Papamarkou, Jacob Hinkle, M.~Todd Young, and David Womble.
\newblock Challenges in {M}arkov chain {M}onte {C}arlo for {B}ayesian neural
  networks.
\newblock \emph{Statistical Science}, 37\penalty0 (3):\penalty0 425--442, 2022.

\bibitem[Pedregosa et~al.(2011)Pedregosa, Varoquaux, Gramfort, Michel, Thirion,
  Grisel, Blondel, Prettenhofer, Weiss, Dubourg, VanderPlas, Passos,
  Cournapeau, Brucher, Perrot, and Duchesnay]{scikit-learn}
F.~Pedregosa, G.~Varoquaux, A.~Gramfort, V.~Michel, B.~Thirion, O.~Grisel,
  M.~Blondel, P.~Prettenhofer, R.~Weiss, V.~Dubourg, J.~VanderPlas, A.~Passos,
  D.~Cournapeau, M.~Brucher, M.~Perrot, and E.~Duchesnay.
\newblock Scikit-learn: machine learning in {P}ython.
\newblock \emph{Journal of Machine Learning Research}, 12:\penalty0 2825--2830,
  2011.

\bibitem[Poyiadjis et~al.(2011)Poyiadjis, Doucet, and Singh]{Poyiadjis2011}
George Poyiadjis, Arnaud Doucet, and Sumeetpal~S. Singh.
\newblock Particle approximations of the score and observed information matrix
  in state space models with application to parameter estimation.
\newblock \emph{Biometrika}, 98\penalty0 (1):\penalty0 65--80, 2011.

\bibitem[Sch\"{o}n et~al.(2011)Sch\"{o}n, Wills, and Ninness]{Schon2011}
Thomas~B. Sch\"{o}n, Adrian Wills, and Brett Ninness.
\newblock System identification of nonlinear state-space models.
\newblock \emph{Automatica}, 47\penalty0 (1):\penalty0 39--49, 2011.

\bibitem[Sharma et~al.(2023)Sharma, Farquhar, Nalisnick, and
  Rainforth]{Sharma2023}
Mrinank Sharma, Sebastian Farquhar, Eric Nalisnick, and Tom Rainforth.
\newblock Do {B}ayesian neural networks need to be fully stochastic?
\newblock In \emph{Proceedings of the 26th International Conference on
  Artificial Intelligence and Statistics}, volume 206, pages 7694--7722. PMLR,
  2023.

\bibitem[Singhal and Wu(1988)]{Singhal1988}
Sharad Singhal and Lance Wu.
\newblock Training multilayer perceptrons with the extended {K}alman algorithm.
\newblock In \emph{Proceedings of Advances in Neural Information Processing
  Systems}, volume~1, pages 133--140. Morgan-Kaufmann, 1988.

\bibitem[Sj{\"o}lund(2023)]{sjolund2023tutorial}
Jens Sj{\"o}lund.
\newblock A tutorial on parametric variational inference.
\newblock \emph{arXiv preprint arXiv:2301.01236}, 2023.

\bibitem[Welling and Teh(2011)]{Welling2011}
Max Welling and Yee~Whye Teh.
\newblock Bayesian learning via stochastic gradient {L}angevin dynamics.
\newblock In \emph{Proceedings of the 28th International Conference on Machine
  Learning}, pages 681--688. ACM, 2011.

\bibitem[Wilson and Izmailov(2020)]{Wilson2020}
Andrew~G. Wilson and Pavel Izmailov.
\newblock Bayesian deep learning and a probabilistic perspective of
  generalization.
\newblock In \emph{Proceedings of Advances in Neural Information Processing
  Systems}, volume~33, pages 4697--4708. Curran Associates, Inc., 2020.

\bibitem[Zhang et~al.(2020)Zhang, Cooper, and Sa]{ZhangRuqi2020}
Ruqi Zhang, A.~Feder Cooper, and Christopher~De Sa.
\newblock {AMAGOLD}: amortized {M}etropolis adjustment for efficient stochastic
  gradient {MCMC}.
\newblock In \emph{Proceedings of the 23rd International Conference on
  Artificial Intelligence and Statistics}, volume 108, pages 2142--2152. PMLR,
  2020.

\end{thebibliography}
\bibliographystyle{plainnat}

\section*{Checklist}

\begin{enumerate}
	
	\item For all models and algorithms presented, check if you include:
	
	\begin{enumerate}
		\item A clear description of the mathematical setting, assumptions, algorithm, and/or model. [Yes] Please see Sections~\ref{sec:feynman-kac-smc} and \ref{sec:scalable-smcs}.
		\item An analysis of the properties and complexity (time, space, sample size) of any algorithm. [Yes] We have discussed the properties and complexities of the algorithms in Sections~\ref{sec:feynman-kac-smc} and \ref{sec:scalable-smcs}. We have also shown the computational time at the end of Section~\ref{sec:mnist}.
		\item (Optional) Anonymized source code, with specification of all dependencies, including external libraries. [Yes] Please see \url{https://github.com/spdes/pbnn}.
		
	\end{enumerate}
	
	\item For any theoretical claim, check if you include:
	
	\begin{enumerate}
		\item Statements of the full set of assumptions of all theoretical results. [Not Applicable] The paper is not theory focused.
		\item Complete proofs of all theoretical results. [Not Applicable] The paper is not theory focused.
		\item Clear explanations of any assumptions. [Not Applicable] The paper is not theory focused.
	\end{enumerate}
	
	\item For all figures and tables that present empirical results, check if you include:
	
	\begin{enumerate}
		\item The code, data, and instructions needed to reproduce the main experimental results (either in the supplemental material or as a URL). [Yes] We have released the code at \url{https://github.com/spdes/pbnn}.
		\item All the training details (e.g., data splits, hyperparameters, how they were chosen). [Yes] This information can be found in the appendix.
		\item A clear definition of the specific measure or statistics and error bars (e.g., with respect to the random seed after running experiments multiple times). [Yes]
		\item A description of the computing infrastructure used. (e.g., type of GPUs, internal cluster, or cloud provider). [Yes] This information can be found in the appendix.
	\end{enumerate}
	
	\item If you are using existing assets (e.g., code, data, models) or curating/releasing new assets, check if you include:
	
	\begin{enumerate}
		\item Citations of the creator If your work uses existing assets. [Yes]
		\item The license information of the assets, if applicable. [Not Applicable]
		\item New assets either in the supplemental material or as a URL, if applicable. [Not Applicable]
		\item Information about consent from data providers/curators. [Not Applicable]
		\item Discussion of sensible content if applicable, e.g., personally identifiable information or offensive content. [Not Applicable]
	\end{enumerate}
	
	\item If you used crowdsourcing or conducted research with human subjects, check if you include:
	
	\begin{enumerate}
		\item The full text of instructions given to participants and screenshots. [Not Applicable]
		\item Descriptions of potential participant risks, with links to Institutional Review Board (IRB) approvals if applicable. [Not Applicable]
		\item The estimated hourly wage paid to participants and the total amount spent on participant compensation. [Not Applicable]
	\end{enumerate}
	
\end{enumerate}

\newpage

\onecolumn

\appendix

\section{Synthetic parameter estimation in Section~\ref{sec:syn-param-est}}
\label{supp:syn-param-est}
The aim of this experiment is to test whether the proposed methods can estimate the model parameters and posterior distributions.
Details of the experimental setup are given as follows.

\paragraph{Optimisation} We use the Adam optimiser with a learning rate $r(i)$ that is exponentially decaying from 0.1 with a speed of 0.96 for 200 epochs, that is, $r(i) = 0.1 \times 0.96^{i \, / \, 10}$.
As a batch size we use 10.
The initial values for $\phi$ and $\psi$ are $\begin{bmatrix}0 & 0\end{bmatrix}$ and $0.1$, respectively.

\paragraph{OHSMC setting} We use a random walk kernel with a step size of 0.001.
We also apply the stratified resampling at every iteration.

\paragraph{SGSMC setting} We apply a Metropolis--Rosenbluth--Teller--Hasting MCMC kernel with a random walk proposal with 10 steps and a step size of 0.001.
Other settings are the same as for OHSMC.

\paragraph{Brute-force Monte Carlo setting} By brute-force Monte Carlo, we mean that we estimate the MLE log-likelihood by a Monte Carlo estimation with 10,000 samples.
More specifically, we use Monte Carlo to compute the lower bound $\int \sum_{n=1}^{100} \log p(y_n \cond \phi; \psi) \, \pi(\phi) \diff \phi$.

\paragraph{Hamiltonian Monte Carlo setting} We use 100 leapfrog steps with a step size of 0.01.
The mass is an identity matrix.
We draw 3,000 samples and discard the first 2,000 burn-in samples. 

\paragraph{Variational Bayes setting} The approximate variational family is a mean-field Gaussian initialized with a zero mean and an identity matrix as the covariance.
We jointly optimise the variational parameters and model parameters with respect to the evidence lower bound (ELBO), by using 100 Monte Carlo samples to approximate the expectation in the ELBO.

\paragraph{SWAG setting} The SWAG method uses the pre-trained results from the MAP estimator, and then performs 100 epochs for approximating the posterior mean and covariance.
For the covariance, we use the low-rank approximation with rank $K=20$.

\begin{figure*}[h!]
	\centering
	\includegraphics[width=.99\linewidth]{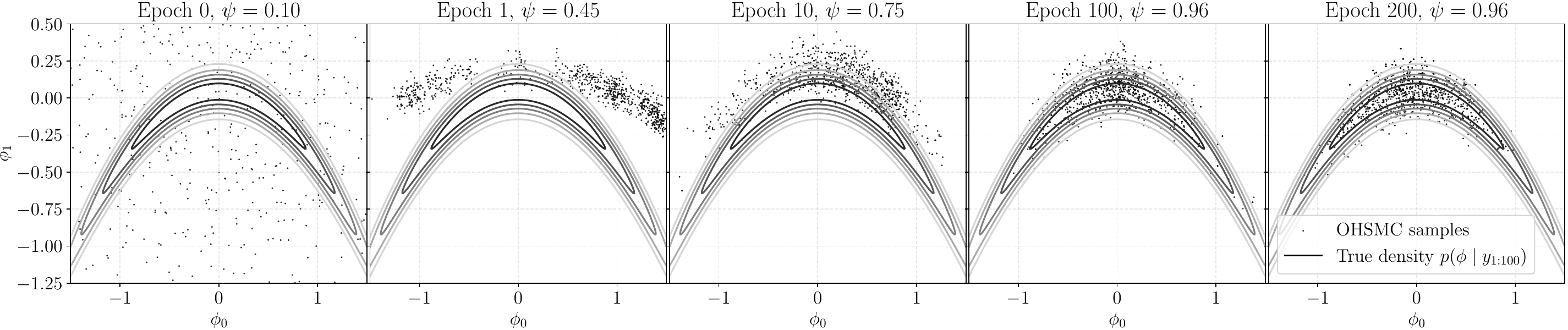}
	\caption{Visualising the flow of the posterior and parameter estimates by OHSMC. In the first epoch, the samples are drawn from the prior. We see that as the epoch increases, the estimates for both $\psi$ and $\phi$ converge.}
	\label{fig:flow}
\end{figure*}

In addition to the results presented in Section~\ref{sec:syn-param-est}, we plot the evolution of the posterior estimation of OHSMC in Figure~\ref{fig:flow}.
There, we can see that for this problem the proposed method converges to the true posterior distribution and parameters with a small bias.

\section{Synthetic regression and classification in Section~\ref{sec:syn-reg-cla}}
\label{supp:syn-reg-cla}
The aim of this section is to profile the methods for training a partial Bayesian neural network (pBNN) for regression and classification.
The details of the experiments are given as follows. 

\paragraph{Regression pBNN} The pBNN configuration is shown in Table~\ref{tbl:syn-reg-nn}. 
\begin{table}[t!]
	\caption{The pBNN configuration for the synthetic regression in Section~\ref{sec:syn-reg-cla}. The second layer is stochastic.}
	\label{tbl:syn-reg-nn}
	\begin{center}
		\begin{tabular}{@{}ccccc@{}}
			\toprule
			Layer & Input & First & Second & Output \\ \midrule
			\begin{tabular}[c]{@{}c@{}} Type (dimension) \\ activation\end{tabular} & (1)   & \begin{tabular}[c]{@{}c@{}}Dense (20)\\ GeLU\end{tabular} & \begin{tabular}[c]{@{}c@{}}Dense (10)\\ GeLU\end{tabular} & \begin{tabular}[c]{@{}c@{}}Dense (1)\\ None\end{tabular}    \\ \bottomrule
		\end{tabular}
	\end{center}
\end{table}

\paragraph{Classification pBNN} The pBNN configuration is shown in Table~\ref{tbl:syn-cla-nn}. 
\begin{table}[t!]
	\caption{The pBNN configuration for the synthetic classification in Section~\ref{sec:syn-reg-cla}. The third layer is stochastic.}
	\label{tbl:syn-cla-nn}
	\begin{center}
		\begin{tabular}{@{}ccclcc@{}}
			\toprule
			Layer & Input & First & Second & Third & Output \\ \midrule
			\begin{tabular}[c]{@{}c@{}}Type (dimension) \\ activation\end{tabular} & (1)   & \begin{tabular}[c]{@{}c@{}}Dense (100)\\ GeLU\end{tabular} & \multicolumn{1}{c}{\begin{tabular}[c]{@{}c@{}}Dense (20)\\ GeLU\end{tabular}} & \begin{tabular}[c]{@{}c@{}}Dense (5)\\ GeLU\end{tabular} & \begin{tabular}[c]{@{}c@{}}Dense (1)\\ Sigmoid\end{tabular} \\ \bottomrule
		\end{tabular}
	\end{center}
\end{table}

\paragraph{Moon data} We create the two-moon data with a function from scikit-learn \citep{scikit-learn} where we set the noise parameter to 0.3.

\paragraph{Optimisation} We use the Adam optimiser with a constant learning rate of 0.01.
In total, we use 100 data points each for training, validation, and testing, respectively.
As a batch size we use 20.
At each batch computation we compute the validation loss and save the best parameters, until we reach a maximum of 200 epochs.
For classification the maximum number of epochs is 100. 

\paragraph{OHSMC setting} The same as in Section~\ref{supp:syn-param-est}, except that we choose the random walk variance to be 0.01.

\paragraph{SGSMC setting} The same as in Section~\ref{supp:syn-param-est}, except that we choose the MCMC random walk variance to be 0.05.
\looseness=-1

\paragraph{Hamiltonian Monte Carlo setting} The same as in Section~\ref{supp:syn-param-est}.

\paragraph{Variational Bayes setting} The same as in Section~\ref{supp:syn-param-est}.

\paragraph{SWAG setting} The same as in Section~\ref{supp:syn-param-est}, but we perform 200 SWAG iterations for approximating the posterior mean and covariance, and we use $K=100$.

\begin{figure}[t!]
	\centering
	\includegraphics[width=.95\linewidth]{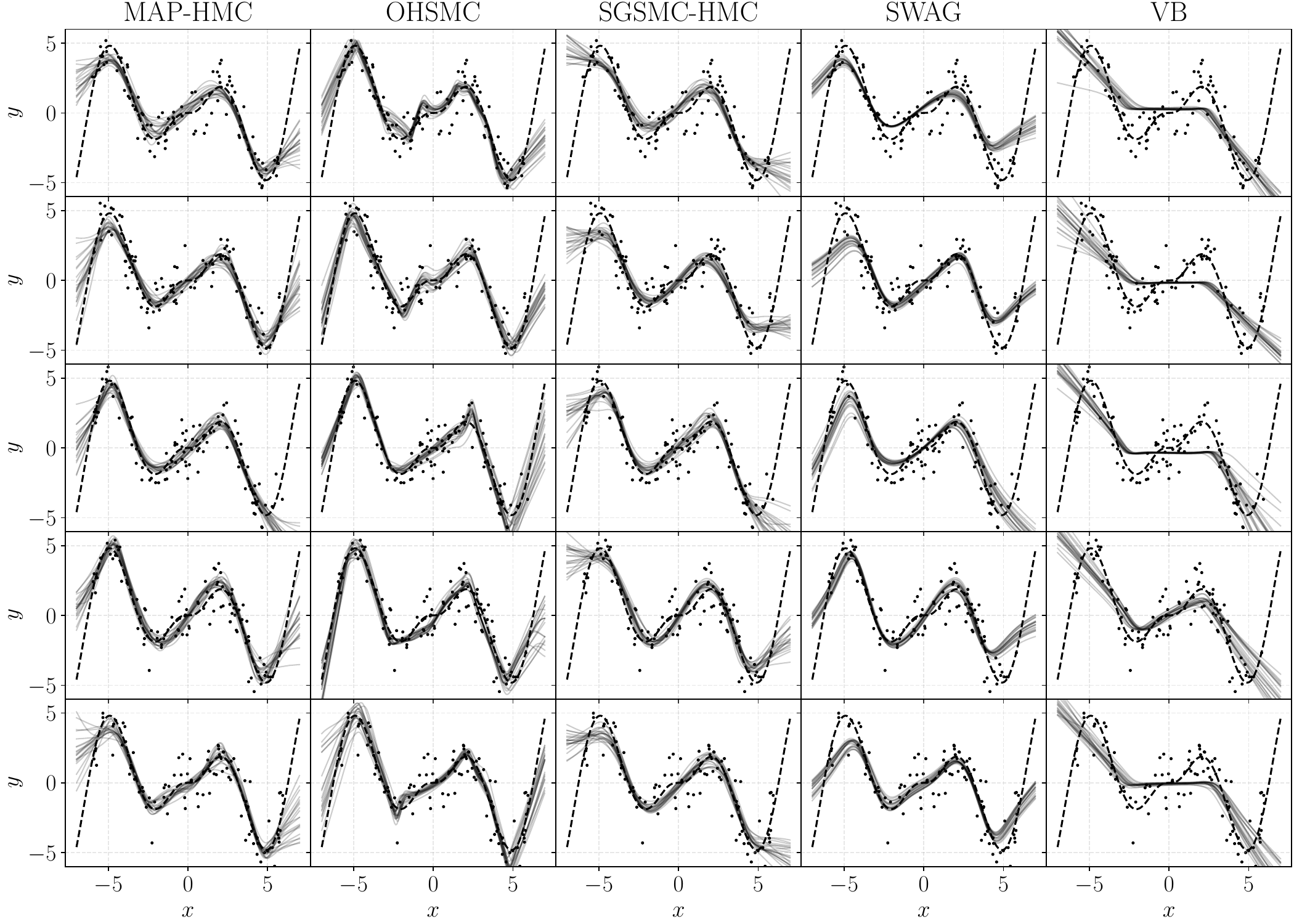}
	\caption{Regression results under five different random seeds. Each row corresponds to another random seed. We see that for all the experiments, OHSMC consistently has the best performance for extrapolation. }
	\label{fig:reg-vis-more}
\end{figure}

\begin{figure}[t!]
	\centering
	\includegraphics[width=.95\linewidth]{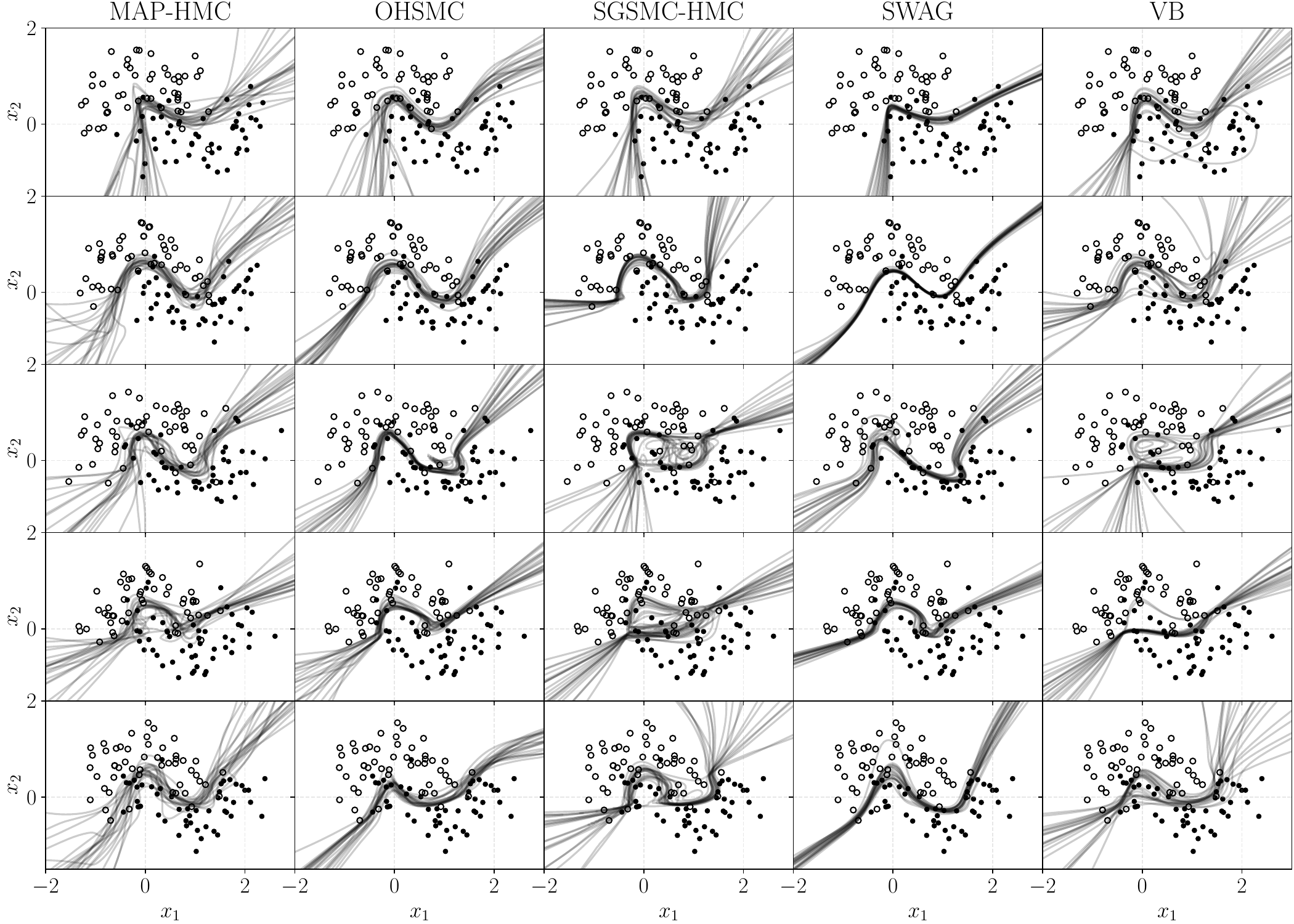}
	\caption{Classification results under five different random seeds. Each row corresponds to another random seed. }
	\label{fig:reg-cla-more}
\end{figure}

Recall that we repeat 100 independent experiments.
In Figures~\ref{fig:reg-vis-more} and~\ref{fig:reg-cla-more}, we additionally plot five regression and classification experiments, respectively.

\section{UCI regression and classification in Section~\ref{sec:uci}}
\label{appendix:uci}
The aim of this experiment is to test whether the proposed methods can train pBNNs for real-world regression and classification tasks.
In addition to what we have described in Section~\ref{sec:uci}, we detail the experimental settings and results as follows. 

\paragraph{Datasets description} In Table~\ref{tbl:uci-data}, we show the details (i.e., number of data points and dimensions) of the UCI datasets \citep{uci} that we use.

\paragraph{pBNN} We use the pBNN defined in Table~\ref{tbl:uci-nn} for UCI regression and classification.
\begin{table}[t!]
	\caption{The pBNN configuration for UCI regression and classification in Section~\ref{sec:uci}. The third layer is stochastic. In the table, $d_x$ and $d_y>1$ stand for the input dimension and number of labels, respectively, depending on the dataset used.}
	\label{tbl:uci-nn}
	\begin{center}
		\begin{tabular}{@{}ccclcc@{}}
			\toprule
			Layer & Input & First & Second & Third & Output \\ \midrule
			\begin{tabular}[c]{@{}c@{}}Type (dimension) \\ activation\end{tabular} & ($d_x$)   & \begin{tabular}[c]{@{}c@{}}Dense (50)\\ GeLU\end{tabular} & \multicolumn{1}{c}{\begin{tabular}[c]{@{}c@{}}Dense (20)\\ GeLU\end{tabular}} & \begin{tabular}[c]{@{}c@{}}Dense (5)\\ GeLU\end{tabular} & \begin{tabular}[c]{@{}c@{}}Dense ($d_y$)\\ Softmax if $d_y>1$ else None\end{tabular} \\ \bottomrule
		\end{tabular}
	\end{center}
\end{table}

\paragraph{Optimisation} The same as in Section~\ref{supp:syn-reg-cla}, except that we apply different batch sizes for each UCI dataset as detailed in Table~\ref{tbl:uci-data}. 

\paragraph{OHSMC setting} We use two Markov kernels.
One is the same random walk kernel as in Section~\ref{supp:syn-reg-cla}.
The other is that of an Ornstein--Uhlenbeck process with the prior as the stationary distribution, and a terminal time of 0.1.
We report the best results under the two Markov kernel choices.

\paragraph{SGSMC setting} The same as in Section~\ref{supp:syn-reg-cla}.

\paragraph{Hamiltonian Monte Carlo setting} The same as in Section~\ref{supp:syn-reg-cla}.

\paragraph{Variational Bayes setting} The same as in Section~\ref{supp:syn-reg-cla}.

\paragraph{SWAG setting} The same as in Section~\ref{supp:syn-reg-cla}, but we use $K=50$.

\paragraph{Detailed results} In Section~\ref{sec:uci}, we have only shown limited results for two regression and two classification datasets.
Table~\ref{tbl:uci-data} depicts more detailed results for seven regression and five classification tasks.
From the table we find that the proposed methods (OHSMC and SGSMC) outperform the other methods on most of the datasets.

\begin{table}[t!]
	\caption{UCI regression and classification results. We uniformly use 60\%, 30\%, and 10\% of the data as training, validation, and test, respectively. Within the tables, $N$, $d_x$, and $d_y$ represent the total data points, input dimensions, and number of classes, respectively. For the regression of naval with SGSMC, the method diverged. Acc. stands for accuracy. }
	\label{tbl:uci-data}
	\begin{center}
		\resizebox{\linewidth}{!}{%
			\begin{tabular}{@{}rcccccccccccccl@{}}
				\toprule
				\multirow{2}{*}{Method} & \multicolumn{2}{c}{boston}                                                                                                      & \multicolumn{2}{c}{concrete}                                                                                                    & \multicolumn{2}{c}{energy}                                                                                                      & \multicolumn{2}{c}{kin8}                                                                                                        & \multicolumn{2}{c}{naval}                                                                                                       & \multicolumn{2}{c}{yacht}                                                                                                       & \multicolumn{2}{c}{power}                                                                                                       \\ \cmidrule(l){2-3} \cmidrule(l){4-5} \cmidrule(l){6-7} \cmidrule(l){8-9} \cmidrule(l){10-11} \cmidrule(l){12-13} \cmidrule(l){14-15} 
				& \multicolumn{1}{r}{NLPD}                                       & \multicolumn{1}{r}{RMSE}                                       & \multicolumn{1}{r}{NLPD}                                       & \multicolumn{1}{r}{RMSE}                                       & \multicolumn{1}{r}{NLPD}                                       & \multicolumn{1}{r}{RMSE}                                       & \multicolumn{1}{r}{NLPD}                                       & \multicolumn{1}{r}{RMSE}                                       & \multicolumn{1}{r}{NLPD}                                       & \multicolumn{1}{r}{RMSE}                                       & \multicolumn{1}{r}{NLPD}                                       & \multicolumn{1}{r}{RMSE}                                       & \multicolumn{1}{r}{NLPD}                                       & RMSE                                                           \\ \midrule
				MAP-HMC                 & \begin{tabular}[c]{@{}c@{}}1.04\\ (0.02)\end{tabular}          & \begin{tabular}[c]{@{}c@{}}0.54\\ (0.09)\end{tabular}          & \begin{tabular}[c]{@{}c@{}}1.00\\ (0.00)\end{tabular}          & \begin{tabular}[c]{@{}c@{}}0.42\\ (0.02)\end{tabular}          & \begin{tabular}[c]{@{}c@{}}0.94\\ (0.00)\end{tabular}          & \begin{tabular}[c]{@{}c@{}}0.24\\ (0.02)\end{tabular}          & \begin{tabular}[c]{@{}c@{}}0.96\\ (0.00)\end{tabular}          & \begin{tabular}[c]{@{}c@{}}0.29\\ (0.00)\end{tabular}          & \begin{tabular}[c]{@{}c@{}}0.94\\ (0.06)\end{tabular}          & \begin{tabular}[c]{@{}c@{}}0.10\\ (0.18)\end{tabular}          & \begin{tabular}[c]{@{}c@{}}0.95\\ (0.01)\end{tabular}          & \begin{tabular}[c]{@{}c@{}}0.27\\ (0.04)\end{tabular}          & \begin{tabular}[c]{@{}c@{}}0.95\\ (0.00)\end{tabular}          & \begin{tabular}[c]{@{}l@{}}0.25\\ (0.01)\end{tabular}          \\
				OHSMC                   & \begin{tabular}[c]{@{}c@{}}\textbf{1.00}\\ (0.02)\end{tabular} & \begin{tabular}[c]{@{}c@{}}\textbf{0.40}\\ (0.08)\end{tabular} & \begin{tabular}[c]{@{}c@{}}\textbf{0.97}\\ (0.01)\end{tabular} & \begin{tabular}[c]{@{}c@{}}\textbf{0.33}\\ (0.03)\end{tabular} & \begin{tabular}[c]{@{}c@{}}\textbf{0.92}\\ (0.00)\end{tabular} & \begin{tabular}[c]{@{}c@{}}\textbf{0.09}\\ (0.00)\end{tabular} & \begin{tabular}[c]{@{}c@{}}\textbf{0.95}\\ (0.00)\end{tabular} & \begin{tabular}[c]{@{}c@{}}\textbf{0.27}\\ (0.00)\end{tabular} & \begin{tabular}[c]{@{}c@{}}0.92\\ (0.00)\end{tabular}          & \begin{tabular}[c]{@{}c@{}}0.05\\ (0.00)\end{tabular}          & \begin{tabular}[c]{@{}c@{}}\textbf{0.92}\\ (0.00)\end{tabular} & \begin{tabular}[c]{@{}c@{}}\textbf{0.11}\\ (0.04)\end{tabular} & \begin{tabular}[c]{@{}c@{}}\textbf{0.94}\\ (0.00)\end{tabular} & \begin{tabular}[c]{@{}l@{}}\textbf{0.24}\\ (0.01)\end{tabular} \\
				SGSMC-HMC                   & \begin{tabular}[c]{@{}c@{}}1.01 \\ (0.05)\end{tabular}         & \begin{tabular}[c]{@{}c@{}}0.42\\ (0.11)\end{tabular}          & \begin{tabular}[c]{@{}c@{}}0.98\\ (0.00)\end{tabular}          & \begin{tabular}[c]{@{}c@{}}0.35\\ (0.02)\end{tabular}          & \begin{tabular}[c]{@{}c@{}}\textbf{0.92}\\ (0.00)\end{tabular} & \begin{tabular}[c]{@{}c@{}}0.14\\ (0.02)\end{tabular}          & \begin{tabular}[c]{@{}c@{}}\textbf{0.95}\\ (0.00)\end{tabular} & \begin{tabular}[c]{@{}c@{}}0.28\\ (0.00)\end{tabular}          & N/A                                                            & N/A                                                            & \begin{tabular}[c]{@{}c@{}}0.93\\ (0.00)\end{tabular}          & \begin{tabular}[c]{@{}c@{}}0.18\\ (0.01)\end{tabular}          & \begin{tabular}[c]{@{}c@{}}\textbf{0.94}\\ (0.00)\end{tabular} & \begin{tabular}[c]{@{}l@{}}\textbf{0.24}\\ (0.01)\end{tabular} \\
				SWAG                    & \begin{tabular}[c]{@{}c@{}}1.02\\ (0.02)\end{tabular}          & \begin{tabular}[c]{@{}c@{}}0.45\\ (0.05)\end{tabular}          & \begin{tabular}[c]{@{}c@{}}0.99\\ (0.01)\end{tabular}          & \begin{tabular}[c]{@{}c@{}}0.39\\ (0.04)\end{tabular}          & \begin{tabular}[c]{@{}c@{}}0.95\\ (0.00)\end{tabular}          & \begin{tabular}[c]{@{}c@{}}0.25\\ (0.02)\end{tabular}          & \begin{tabular}[c]{@{}c@{}}1.10\\ (0.20)\end{tabular}          & \begin{tabular}[c]{@{}c@{}}0.54\\ (0.27)\end{tabular}          & \begin{tabular}[c]{@{}c@{}}1.05\\ (0.20)\end{tabular}          & \begin{tabular}[c]{@{}c@{}}0.46\\ (0.36)\end{tabular}          & \begin{tabular}[c]{@{}c@{}}0.95\\ (0.02)\end{tabular}          & \begin{tabular}[c]{@{}c@{}}0.25\\ (0.08)\end{tabular}          & \begin{tabular}[c]{@{}c@{}}0.96\\ (0.01)\end{tabular}          & \begin{tabular}[c]{@{}l@{}}0.29\\ (0.04)\end{tabular}          \\
				VB                      & \begin{tabular}[c]{@{}c@{}}1.01\\ (0.03)\end{tabular}          & \begin{tabular}[c]{@{}c@{}}0.43\\ (0.07)\end{tabular}          & \begin{tabular}[c]{@{}c@{}}\textbf{0.97}\\ (0.00)\end{tabular} & \begin{tabular}[c]{@{}c@{}}0.34\\ (0.01)\end{tabular}          & \begin{tabular}[c]{@{}c@{}}\textbf{0.93}\\ (0.00)\end{tabular} & \begin{tabular}[c]{@{}c@{}}0.19\\ (0.03)\end{tabular}          & \begin{tabular}[c]{@{}c@{}}\textbf{0.95}\\ (0.00)\end{tabular} & \begin{tabular}[c]{@{}c@{}}\textbf{0.27}\\ (0.00)\end{tabular} & \begin{tabular}[c]{@{}c@{}}\textbf{0.91}\\ (0.00)\end{tabular} & \begin{tabular}[c]{@{}c@{}}\textbf{0.04}\\ (0.00)\end{tabular} & \begin{tabular}[c]{@{}c@{}}0.94\\ (0.00)\end{tabular}          & \begin{tabular}[c]{@{}c@{}}0.20\\ (0.02)\end{tabular}          & \begin{tabular}[c]{@{}c@{}}\textbf{0.94}\\ (0.00)\end{tabular} & \begin{tabular}[c]{@{}l@{}}\textbf{0.24}\\ (0.01)\end{tabular} \\ \midrule
				$(N, d_x)$              & \multicolumn{2}{c}{(506, 13)}                                                                                                   & \multicolumn{2}{c}{(1030, 8)}                                                                                                   & \multicolumn{2}{c}{(768, 8)}                                                                                                    & \multicolumn{2}{c}{(8192, 8)}                                                                                                   & \multicolumn{2}{c}{(11934, 14)}                                                                                                 & \multicolumn{2}{c}{(308, 6)}                                                                                                    & \multicolumn{2}{c}{(9568, 4)}                                                                                                   \\
				Batch size              & \multicolumn{2}{c}{50}                                                                                                          & \multicolumn{2}{c}{50}                                                                                                          & \multicolumn{2}{c}{50}                                                                                                          & \multicolumn{2}{c}{50}                                                                                                          & \multicolumn{2}{c}{50}                                                                                                          & \multicolumn{2}{c}{20}                                                                                                          & \multicolumn{2}{c}{50}                                                                                                          \\ \bottomrule
			\end{tabular}
		}
		\newline
		\resizebox{\linewidth}{!}{%
			\begin{tabular}{@{}rccccccccccccccc@{}}
				\toprule
				\multirow{2}{*}{Method} & \multicolumn{3}{c}{australian}                                                                                                                                                                      & \multicolumn{3}{c}{cancer}                                                                                                                                                                          & \multicolumn{3}{c}{ionosphere}                                                                                                                                                                      & \multicolumn{3}{c}{glass}                                                                                                                                                                           & \multicolumn{3}{c}{satellite}                                                                                                                                                                       \\ \cmidrule(l){2-4} \cmidrule(l){5-7} \cmidrule(l){8-10} \cmidrule(l){11-13} \cmidrule(l){14-16}
				& \multicolumn{1}{r}{NLPD}                                        & \multicolumn{1}{r}{ECE}                                         & \multicolumn{1}{r}{Acc.}                                         & \multicolumn{1}{r}{NLPD}                                        & \multicolumn{1}{r}{ECE}                                         & \multicolumn{1}{r}{Acc.}                                         & \multicolumn{1}{r}{NLPD}                                        & \multicolumn{1}{r}{ECE}                                         & \multicolumn{1}{r}{Acc.}                                         & \multicolumn{1}{r}{NLPD}                                        & \multicolumn{1}{r}{ECE}                                         & \multicolumn{1}{r}{Acc.}                                         & \multicolumn{1}{r}{NLPD}                                        & \multicolumn{1}{r}{ECE}                                         & \multicolumn{1}{r}{Acc.}                                         \\ \midrule
				MAP-HMC                 & \begin{tabular}[c]{@{}c@{}}0.36 \\ (0.10)\end{tabular}          & \begin{tabular}[c]{@{}c@{}}\textbf{0.09} \\ (0.02)\end{tabular} & \begin{tabular}[c]{@{}c@{}}0.85 \\ (0.03)\end{tabular}          & \begin{tabular}[c]{@{}c@{}}0.12 \\ (0.06)\end{tabular}          & \begin{tabular}[c]{@{}c@{}}0.04 \\ (0.01)\end{tabular}          & \begin{tabular}[c]{@{}c@{}}0.95 \\ (0.02)\end{tabular}          & \begin{tabular}[c]{@{}c@{}}\textbf{0.21} \\ (0.09)\end{tabular} & \begin{tabular}[c]{@{}c@{}}0.09 \\ (0.03)\end{tabular}          & \begin{tabular}[c]{@{}c@{}}0.90 \\ (0.04)\end{tabular}          & \begin{tabular}[c]{@{}c@{}}1.14 \\ (0.16)\end{tabular}          & \begin{tabular}[c]{@{}c@{}}0.27 \\ (0.06)\end{tabular}          & \begin{tabular}[c]{@{}c@{}}0.58 \\ (0.08)\end{tabular}          & \begin{tabular}[c]{@{}c@{}}0.28 \\ (0.03)\end{tabular}          & \begin{tabular}[c]{@{}c@{}}0.04 \\ (0.01)\end{tabular}          & \begin{tabular}[c]{@{}c@{}}\textbf{0.90}\\ (0.00)\end{tabular}  \\
				OHSMC                   & \begin{tabular}[c]{@{}c@{}}0.37 \\ (0.09)\end{tabular}          & \begin{tabular}[c]{@{}c@{}}\textbf{0.09} \\ (0.02)\end{tabular} & \begin{tabular}[c]{@{}c@{}}0.85 \\ (0.03)\end{tabular}          & \begin{tabular}[c]{@{}c@{}}0.14 \\ (0.08)\end{tabular}          & \begin{tabular}[c]{@{}c@{}}\textbf{0.03} \\ (0.01)\end{tabular} & \begin{tabular}[c]{@{}c@{}}\textbf{0.96} \\ (0.02)\end{tabular} & \begin{tabular}[c]{@{}c@{}}\textbf{0.21} \\ (0.13)\end{tabular} & \begin{tabular}[c]{@{}c@{}}\textbf{0.08} \\ (0.04)\end{tabular} & \begin{tabular}[c]{@{}c@{}}0.90 \\ (0.04)\end{tabular}          & \begin{tabular}[c]{@{}c@{}}\textbf{1.13} \\ (0.32)\end{tabular} & \begin{tabular}[c]{@{}c@{}}0.29 \\ (0.06)\end{tabular}          & \begin{tabular}[c]{@{}c@{}}\textbf{0.62} \\ (0.06)\end{tabular} & \begin{tabular}[c]{@{}c@{}}\textbf{0.27} \\ (0.03)\end{tabular} & \begin{tabular}[c]{@{}c@{}}\textbf{0.03} \\ (0.01)\end{tabular} & \begin{tabular}[c]{@{}c@{}}0.89 \\ (0.01)\end{tabular}          \\
				SGSMC-HMC                   & \begin{tabular}[c]{@{}c@{}}0.36 \\ (0.11)\end{tabular}          & \begin{tabular}[c]{@{}c@{}}\textbf{0.09} \\ (0.03)\end{tabular} & \begin{tabular}[c]{@{}c@{}}0.85 \\ (0.04)\end{tabular}          & \begin{tabular}[c]{@{}c@{}}0.12 \\ (0.09)\end{tabular}          & \begin{tabular}[c]{@{}c@{}}\textbf{0.03} \\ (0.02)\end{tabular} & \begin{tabular}[c]{@{}c@{}}\textbf{0.96} \\ (0.02)\end{tabular} & \begin{tabular}[c]{@{}c@{}}0.24 \\ (0.19)\end{tabular}          & \begin{tabular}[c]{@{}c@{}}\textbf{0.08} \\ (0.04)\end{tabular} & \begin{tabular}[c]{@{}c@{}}\textbf{0.92} \\ (0.05)\end{tabular} & \begin{tabular}[c]{@{}c@{}}1.25 \\ (0.32)\end{tabular}          & \begin{tabular}[c]{@{}c@{}}0.26 \\ (0.07)\end{tabular}          & \begin{tabular}[c]{@{}c@{}}0.56 \\ (0.10)\end{tabular}          & \begin{tabular}[c]{@{}c@{}}0.84 \\ (0.74)\end{tabular}          & \begin{tabular}[c]{@{}c@{}}0.10 \\ (0.10)\end{tabular}          & \begin{tabular}[c]{@{}c@{}}0.79 \\ (0.14)\end{tabular}          \\
				SWAG                    & \begin{tabular}[c]{@{}c@{}}0.36 \\ (0.08)\end{tabular}          & \begin{tabular}[c]{@{}c@{}}\textbf{0.09} \\ (0.02)\end{tabular} & \begin{tabular}[c]{@{}c@{}}\textbf{0.86} \\ (0.03)\end{tabular} & \begin{tabular}[c]{@{}c@{}}0.14 \\ (0.04)\end{tabular}          & \begin{tabular}[c]{@{}c@{}}0.06 \\ (0.03)\end{tabular}          & \begin{tabular}[c]{@{}c@{}}0.95 \\ (0.02)\end{tabular}          & \begin{tabular}[c]{@{}c@{}}0.36 \\ (0.12)\end{tabular}          & \begin{tabular}[c]{@{}c@{}}0.14 \\ (0.05)\end{tabular}          & \begin{tabular}[c]{@{}c@{}}0.84 \\ (0.08)\end{tabular}          & \begin{tabular}[c]{@{}c@{}}1.36 \\ (0.41)\end{tabular}          & \begin{tabular}[c]{@{}c@{}}0.26 \\ (0.06)\end{tabular}          & \begin{tabular}[c]{@{}c@{}}0.58 \\ (0.10)\end{tabular}          & \begin{tabular}[c]{@{}c@{}}0.35 \\ (0.07)\end{tabular}          & \begin{tabular}[c]{@{}c@{}}0.07 \\ (0.04)\end{tabular}          & \begin{tabular}[c]{@{}c@{}}0.85 \\ (0.03)\end{tabular}          \\
				VB                      & \begin{tabular}[c]{@{}c@{}}\textbf{0.34} \\ (0.07)\end{tabular} & \begin{tabular}[c]{@{}c@{}}0.10 \\ (0.04)\end{tabular}          & \begin{tabular}[c]{@{}c@{}}0.83 \\ (0.03)\end{tabular}          & \begin{tabular}[c]{@{}c@{}}\textbf{0.11} \\ (0.08)\end{tabular} & \begin{tabular}[c]{@{}c@{}}0.04 \\ (0.03)\end{tabular}          & \begin{tabular}[c]{@{}c@{}}\textbf{0.96} \\ (0.02)\end{tabular} & \begin{tabular}[c]{@{}c@{}}0.28 \\ (0.16)\end{tabular}          & \begin{tabular}[c]{@{}c@{}}0.10 \\ (0.05)\end{tabular}          & \begin{tabular}[c]{@{}c@{}}0.89 \\ (0.05)\end{tabular}          & \begin{tabular}[c]{@{}c@{}}1.79 \\ (1.89)\end{tabular}          & \begin{tabular}[c]{@{}c@{}}\textbf{0.22} \\ (0.09)\end{tabular} & \begin{tabular}[c]{@{}c@{}}0.51 \\ (0.11)\end{tabular}          & \begin{tabular}[c]{@{}c@{}}0.28 \\ (0.04)\end{tabular}          & \begin{tabular}[c]{@{}c@{}}\textbf{0.03} \\ (0.00)\end{tabular} & \begin{tabular}[c]{@{}c@{}}\textbf{0.90} \\ (0.00)\end{tabular} \\ \midrule
				$(N, d_x, d_y)$         & \multicolumn{3}{c}{(690, 14, 2)}                                                                                                                                                                    & \multicolumn{3}{c}{(569, 30, 2)}                                                                                                                                                                    & \multicolumn{3}{c}{(351, 33, 2)}                                                                                                                                                                    & \multicolumn{3}{c}{(214, 9, 6)}                                                                                                                                                                     & \multicolumn{3}{c}{(6435, 36, 6)}                                                                                                                                                                   \\
				Batch size              & \multicolumn{3}{c}{50}                                                                                                                                                                              & \multicolumn{3}{c}{50}                                                                                                                                                                              & \multicolumn{3}{c}{20}                                                                                                                                                                              & \multicolumn{3}{c}{20}                                                                                                                                                                              & \multicolumn{3}{c}{50}                                                                                                                                                                              \\ \bottomrule
			\end{tabular}
		}
		\newline
	\end{center}
\end{table}

\section{MNIST classification in Section~\ref{sec:mnist}}
\label{supp:mnist}
The aim of this experiment is to test whether the proposed methods can be applied on the commonly used MNIST dataset \citep{lecun1998mnist}, and to see whether the proposed methods outperform the baselines.
For this experiment, the MCMC-based methods no longer apply due to their demanding computation.
Detailed experiment settings are given in the following.

\paragraph{pBNN setting} The pBNN configuration is given in Table~\ref{tbl:mnist-pbnn}.

\begin{table}[t!]
	\caption{The pBNN for MNIST. The stochastic layer is the first convolution layer.}
	\label{tbl:mnist-pbnn}
	\begin{center}
		\resizebox{\linewidth}{!}{%
		\begin{tabular}{@{}cccccc@{}}
			\toprule
			Layer                    & Input       & First                          & Second                         & Third       & Output     \\ \midrule
			Type (dimension)         & (28, 28, 1) & Conv                           & Conv                           & Dense (256) & Dense (10) \\
			Convolution (size)       & None        & 32 filters (3, 3)              & 64 filters (3, 3)              & None        & None       \\
			Activation               & None        & ReLU                           & ReLU                           & ReLU        & Softmax    \\
			Pooling (size), (stride) & None        & Average pooling (2, 2), (2, 2) & Average pooling (2, 2), (2, 2) & None        & None       \\ \bottomrule
		\end{tabular}
		}
	\end{center}
\end{table}

\paragraph{Optimisation} We use an an Adam optimiser with a constant learning rate of 0.002.
In total, we use 50,000 training, 10,000 validation, and 10,000 test data points.
As a batch size we use 100.
At each batch computation we compute the validation loss and save the best parameters, until we reach a maximum of 10 epochs.
It is worth remarking that the VB method failed to converge for learning rate greater than 0.002.

\paragraph{OHSMC setting} The same as in Section~\ref{supp:syn-reg-cla}.

\paragraph{SGSMC setting} The same as in Section~\ref{supp:syn-reg-cla}.

\paragraph{Hamiltonian Monte Carlo setting} The same as in Section~\ref{supp:syn-reg-cla}.

\paragraph{Variational Bayes setting} The same as in Section~\ref{supp:syn-reg-cla}.

\paragraph{SWAG setting} The same as in Section~\ref{supp:syn-reg-cla}, but we use $K=100$ and perform 50 SWAG iterations for the Gaussian approximation. 

\section{Implementation and computing infrastructure}

All experiments are implemented in Python using JAX,  Optax~\citep{jax2018github}, BlackJAX \citep{blackjax2020github}, and Flax~\citep{flax2020github}.
We run the experiments in the computational cluster Berzelius provided by National Supercomputer Centre at Link\"{o}ping University, and we uniformly use one NVIDIA A100 40 GB GPU for all the experiments.

\end{document}